\documentclass[10pt,journal,compsoc]{IEEEtran}
\usepackage{bm}
\usepackage{multirow, makecell}
\usepackage{algorithm,algorithmicx,algpseudocode}
\algnewcommand{\algorithmicforeach}{\textbf{for each}}
\algdef{SE}[FOR]{ForEach}{EndForEach}[1]
  {\algorithmicforeach\ #1\ \algorithmicdo}% \ForEach{#1}
  {\algorithmicend\ \algorithmicforeach}% \EndForEach

\usepackage{caption}
\usepackage{subcaption}

\usepackage{times}
\usepackage{epsfig}
\usepackage{graphicx}
\usepackage{amsmath}
\usepackage{amssymb}
\usepackage{booktabs}
\usepackage{tabu}
\usepackage[dvipsnames]{xcolor}
\usepackage{spreadtab}
\usepackage{tabularx}

\DeclareMathOperator*{\argmin}{arg\,min}
\usepackage{hyperref}
\usepackage{amsthm}

\theoremstyle{definition}

\usepackage{tikz}
\newcommand*\circled[1]{\tikz[baseline=(char.base)]{
            \node[shape=circle,draw,inner sep=2pt] (char) {#1};}}

\ifCLASSOPTIONcompsoc
  \usepackage[nocompress]{cite}
\else
  \usepackage{cite}
\fi

\definecolor{RoyalBlue}{RGB}{65, 105, 225} % Defining Royal Blue
\definecolor{Coral}{RGB}{255, 127, 80}     % Defining Coral

\ifCLASSINFOpdf
\else
\fi

\begin{document}

\title{CPPF++: Uncertainty-Aware Sim2Real Object Pose Estimation by Vote Aggregation}

% \author{Yang You, Wenhao He, Michael Xu LIU, Weiming Wang\thanks{Cewu Lu and Weiming Wang are the corresponding authors. Cewu Lu is member of Qing Yuan Research Institute and MoE Key Lab of Artificial Intelligence, AI Institute, Shanghai Jiao Tong University, China and Shanghai Qi Zhi institute.}\ , Cewu Lu\footnotemark[1] \\ 
% Shanghai Jiao Tong University, China\\
% \{qq456cvb, eliphat, wangweiming, lucewu\}@sjtu.edu.cn %
% }
\author{Yang You\textsuperscript{1,3}, \hspace{.1cm} Wenhao He\textsuperscript{2}, \hspace{.1cm} Jin Liu\textsuperscript{1}, \hspace{.1cm} \\Hongkai Xiong\textsuperscript{1}, \hspace{.1cm} Weiming Wang\textsuperscript{1}, \hspace{.1cm} Cewu Lu\textsuperscript{1}~\IEEEmembership{Member,~IEEE} \\
\textsuperscript{1}Shanghai Jiao Tong University, \hspace{.1cm}
\textsuperscript{2}Flexiv Robotics Inc. \hspace{.1cm}\\
\textsuperscript{3}Stanford University\hspace{.1cm}
%{\tt\small \textsuperscript{1}ankan.bhunia@mbzuai.ac.ae }
\IEEEcompsocitemizethanks{\IEEEcompsocthanksitem Work done at Shanghai Jiao Tong University. Yang You is with Stanford University, Stanford, United States. Jin Liu, Hongkai Xiong, Weiming Wang, Cewu Lu are with Shanghai Jiao Tong University, Shanghai, China. Wenhao He is with Flexiv Robotics Inc., Shanghai, China.  Cewu Lu is also the member of Shanghai Artificial Intelligence Laboratory (National Laboratory), Shanghai, China.
\\
\IEEEcompsocthanksitem Weiming Wang and Cewu Lu are the corresponding authors.\protect\\
E-mail: wangweiming@sjtu.edu.cn, lucewu@sjtu.edu.cn.
}

}

\IEEEtitleabstractindextext{%
\begin{abstract}
Object pose estimation constitutes a critical area within the domain of 3D vision. While contemporary state-of-the-art methods that leverage real-world pose annotations have demonstrated commendable performance, the procurement of such real training data incurs substantial costs. This paper focuses on a specific setting wherein only 3D CAD models are utilized as a priori knowledge, devoid of any background or clutter information. We introduce a novel method, CPPF++, designed for sim-to-real pose estimation. This method builds upon the foundational point-pair voting scheme of CPPF, reformulating it through a probabilistic view. To address the challenge posed by vote collision, we propose a novel approach that involves modeling the voting uncertainty by estimating the probabilistic distribution of each point pair within the canonical space. Furthermore, we augment the contextual information provided by each voting unit through the introduction of $N$-point tuples. To enhance the robustness and accuracy of the model, we incorporate several innovative modules, including noisy pair filtering, online alignment optimization, and a tuple feature ensemble. Alongside these methodological advancements, we introduce a new category-level pose estimation dataset, named DiversePose 300.
Empirical evidence demonstrates that our method significantly surpasses previous sim-to-real approaches and achieves comparable or superior performance on novel datasets. Our code is available on \href{https://github.com/qq456cvb/CPPF2}{https://github.com/qq456cvb/CPPF2}.
\end{abstract}

}

\maketitle

\IEEEdisplaynontitleabstractindextext

%%%%%%%%% BODY TEXT
\IEEEpeerreviewmaketitle

\ifCLASSOPTIONcompsoc
\IEEEraisesectionheading{\section{Introduction}\label{sec:introduction}}
\else
\section{Introduction}
\label{sec:introduction}
\fi

\label{sec:intro}

Object pose estimation is a pivotal subject within the field of computer vision, with applications extending to various downstream tasks such as robotic manipulation~\cite{stevvsic2020learning,deng2020self} and augmented reality (AR) instructions~\cite{su2019deep}. The exploration of both instance-level and category-level pose estimation methods has been undertaken. Instance-level methods necessitate the availability of the exact object of interest, whereas category-level methods utilize a collection of similar objects (i.e., belonging to the same category) during training. Predominantly, state-of-the-art algorithms in both settings require training on real-world annotations, which are not only costly to acquire but also lack generalizability to out-of-distribution scenarios.

Some researchers have sought to mitigate these challenges by synthetically rendering CAD models with simulated~\cite{hodavn2020bop,zhong2022sim2real,xiang2018posecnn} or real-world~\cite{wang2019normalized} backgrounds for training. While this approach alleviates the need for real-world annotations, it introduces new complications, such as the complexity of generating random backgrounds and placing disturbers and the prior assumption of the background distribution which may not hold in general.

\begin{figure}[t]
\centering
\includegraphics[width=\linewidth]{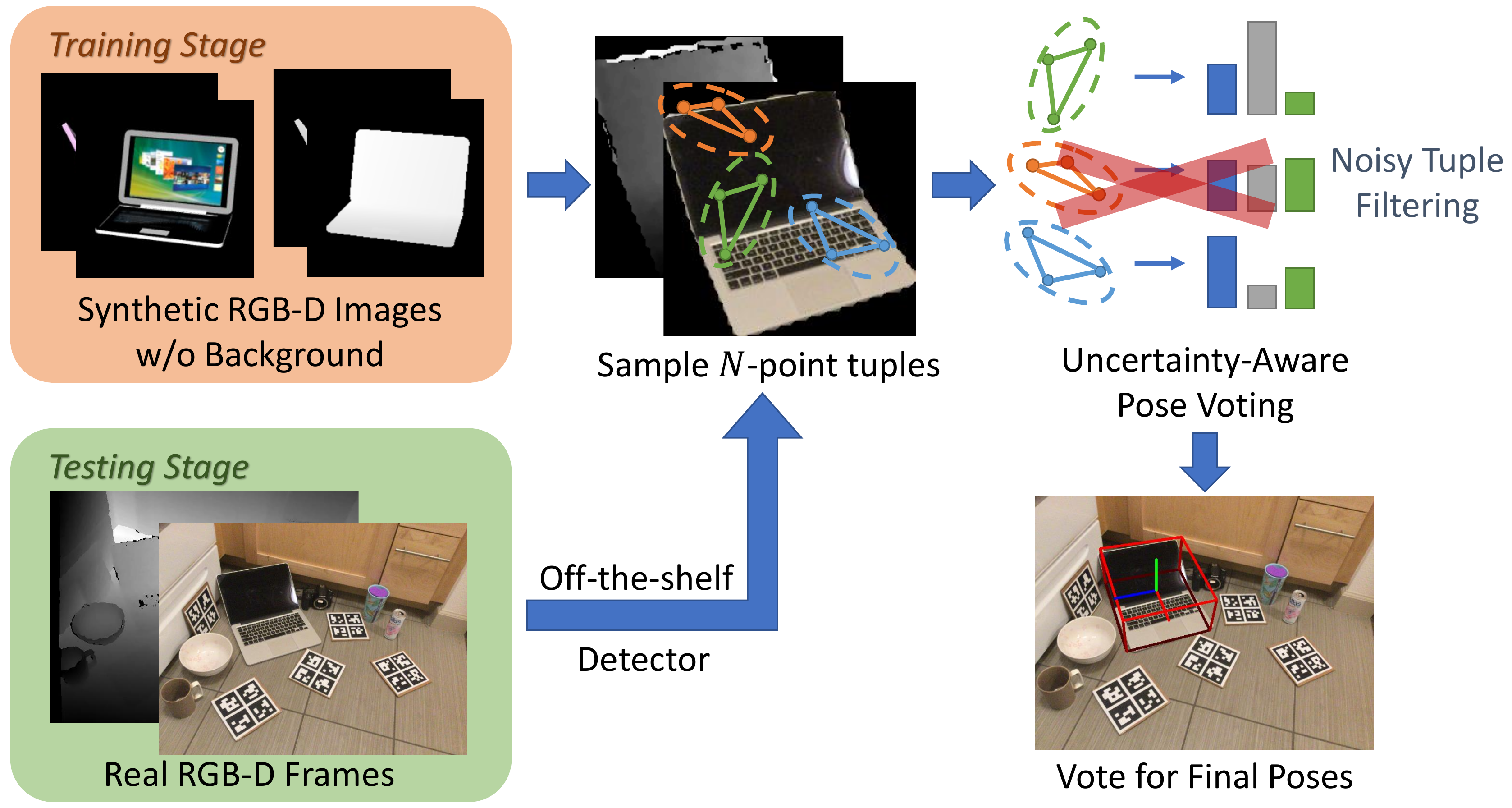}
\caption{Unlike previous methods, our method leverages synthetic RGB-D images without backgrounds for training. During inference, a collection of $N$-point tuples are uniformly sampled to vote poses with uncertainty awareness, culminating in the final prediction as the majority vote.}
\label{fig:intro}
\end{figure}

In this paper, we delve into the setting of pure sim-to-real training, where 3D objects are placed and rendered without any background during training. We contend that this setting is both practical and realistic, as it aligns with common scenarios where only a collection of scanned or synthetic CAD models are available, and rapid training of a pose detection model is desired without prior knowledge of potential backgrounds or clutters.

Building on the recent work of CPPF~\cite{you2022cppf}, which explored this problem setting and introduced a sim-to-real approach, we propose an enhanced voting method called CPPF++. Our method absorbs the voting scheme of CPPF but reformulates it from a probabilistic perspective. We model input point pairs as a multinomial distribution in the canonical space, sampling it to generate votes, and employ noisy pair filtering to mitigate background noise. Furthermore, we introduce $N$-point tuples to preserve more context information and present three rotation-invariant features to maintain rotation invariance. During inference time, to further improve the model's performance we propose a novel online alignment optimization module to refine the output pose differentiably.

We evaluate the performance of current state-of-the-art methods on four different pose estimation datasets including NOCS REAL275, Wild6D, PhoCAL and DiversePose 300. Among these, DiversePose 300, our newly proposed dataset, presents a significant challenge within category-level pose estimation. It offers an expansive variety of poses and background settings, aiming to augment the scope of existing pose estimation datasets. The experiments reveal that our method substantially surpasses prior sim-to-real techniques across all datasets. Furthermore, it's noteworthy that while state-of-the-art real-world training methods might exhibit overfitting on NOCS REAL275, they fall short when compared to our method on the other unseen datasets.

In summary, our contributions are three-fold:
\begin{itemize}
\item We approach the pose voting process from a probabilistic perspective, estimating the distribution of canonical coordinates of point pairs to manage uncertainty during voting. Our framework is bolstered by several innovative modules, including noisy pair filtering, online alignment optimization, and a tuple feature ensemble, all designed to enhance performance.
\item Our experimental results demonstrate substantial improvements over existing sim-to-real techniques on the NOCS REAL275~\cite{wang2019normalized} dataset for category-level pose estimation. Moreover, our method exhibits superior generalizability to unseen datasets (e.g., Wild6D, PhoCAL, and DiversePose 300), outperforming previous approaches. Additionally, we achieve enhancements over traditional Point Pair Features (PPF) in instance-level pose estimation on the YCB-Video~\cite{xiang2018posecnn} dataset.
\item We present the DiversePose 300 dataset, a more challenging category-level pose dataset that emphasizes a wide variety of poses and background distributions. This dataset serves as a complementary resource to existing pose estimation datasets, aiming to address their limitations.
\end{itemize}
%-------------------------------------------------------------------------
\section{Related Works}

\subsection{Pose Estimation Trained on Real-World Data}
Currently, quite a lot of methods have been proposed for 6D pose estimation. 
For instance-level methods, PoseCNN~\cite{xiang2018posecnn} regresses the depth and 2D center offsets and a quaternion rotation for each region-of-interest with 2D CNNs. DeepIM~\cite{li2018deepim} proposes to iteratively refine the initial pose estimation by giving a pose residual of comparing the rendered image and the input image. CosyPose~\cite{labbe2020cosypose} improves upon DeepIM by matching individual 6D object pose hypotheses across different images in order to jointly estimate camera viewpoints and 6D poses consistently. 
DenseFusion~\cite{wang2019densefusion} learns a per-point confidence in RGB-D point clouds, and predicts the final pose with the best response. RCVPose~\cite{wu2022vote} votes for keypoint locations by finding intersections of multiple spheres, and then use PnP to solve the final pose.
% PVNet~\cite{peng2019pvnet} regresses the vector fields toward the 2D projections of the chosen 3D key points and uses weighted PnP to estimate the final location. 
In addition to instance-level methods, NOCS~\cite{wang2019normalized} first introduces a category-level pose estimation dataset and a regression method to learn the normalized coordinates of an object. CASS~\cite{chen2020learning} proposes a variational auto-encoder to capture pose-independent features, along with pose dependent ones to predict 6D poses. DualPoseNet~\cite{Lin_2021_ICCV} uses two parallel decoders to make both an explicit and implicit representation of an object's 6D pose. CenterSnap~\cite{irshad2022centersnap} represents object instances as center keypoints in a spatial 2D grid, and then regresses the object 6D pose at each position. ShAPO~\cite{irshad2022shapo} improves CenterSnap by proposing a differentiable pipeline to improve the initial pose, along with the shape and appearance, using an octree structure. 
% There are many other methods~\cite{shi2021stablepose,kehl2017ssd,brachmann2014learning} that focus on instance-level pose estimation.
The above methods are mostly end-to-end and require real-world training data to work well.

Another research direction concentrates on training with real-world data in the absence of ground-truth annotations, as seen in works like RePoNet~\cite{ze2022category}, UDA-COPE~\cite{lee2022uda}, and TTA-COPE~\cite{lee2023tta}. However, these approaches assume a prior distribution of background noise in the target test domain, rendering them less robust when encountering novel scenes.

\subsection{Pose Estimation Trained on Synthetic Data}
Fewer methods have been proposed to handle the problem of sim-to-real pose estimation. For instance-level pose estimation, Zhong \textit{et al.}~\cite{zhong2022sim2real} develop a sim-to-real contrastive learning mechanism that can generalize the model trained in simulation to real-world applications. Likewise, TemplatePose~\cite{nguyen2022templates} uses local feature encoder to compare the query image and the pre-rendered images from different views. It can be trained once and directly generalize to new objects. SurfEmb~\cite{haugaard2022surfemb} learns dense and continuous 2D-3D correspondence distributions between the query image and the target model. It uses physical-based rendered training data provided by BOP challenge~\cite{hodavn2020bop}, which are generated offline. For category-level pose estimation, Chen \textit{et al.}~\cite{chen2020category} propose to render synthetic models and compare the appearance with real images in different poses. Perhaps, CPPF~\cite{you2022cppf} is the closest to this paper. It designs a special voting scheme by sampling numerous point pairs and predicting the voting targets. It is robust to noise and clutter and uses only the rendered synthetic object itself. However, its performance is still not as good as the current state-of-the-art, and it only conducts experiments on category-level pose estimation.

\section{Method}

In this section, we introduce the structure and key components of our proposed methodology. We begin by examining the most closely related work, CPPF, as detailed in Section~\ref{sec:pre}.

Our work, denoted as CPPF++, is an extension of CPPF, and we introduce the following novel and critical strategies to enhance performance:

\begin{itemize}
    \item \textbf{Vote Uncertainty Modeling:} This strategy is elaborated in Section~\ref{sec:probabilistic}, where we discuss the incorporation of uncertainty into the model to improve robustness.
    \item \textbf{N-Point Tuple Feature Extraction:} Section~\ref{sec:npoint} is dedicated to our introduction of N-point tuples, a higher-order feature representation that captures more complex relationships within the data.
    \item \textbf{Noisy Pair Filtering:} In Section~\ref{sec:iterative}, we describe our approach to filter out noisy pairs, a technique designed to refine the model's rotation prediction by reducing noise.
    \item \textbf{Online Alignment Optimization:}  As discussed in Section~\ref{sec:opt}, this process involves differentiating the coordinate alignment loss and refining the output pose for improved model performance.
    \item \textbf{Tuple Feature Ensemable} In Section~\ref{sec:ensemble}, we advocate for the amalgamation of geometric and visual features through an innovative inference-time model switching strategy.
\end{itemize}

\begin{figure*}
    \centering
    \includegraphics[width=\linewidth]{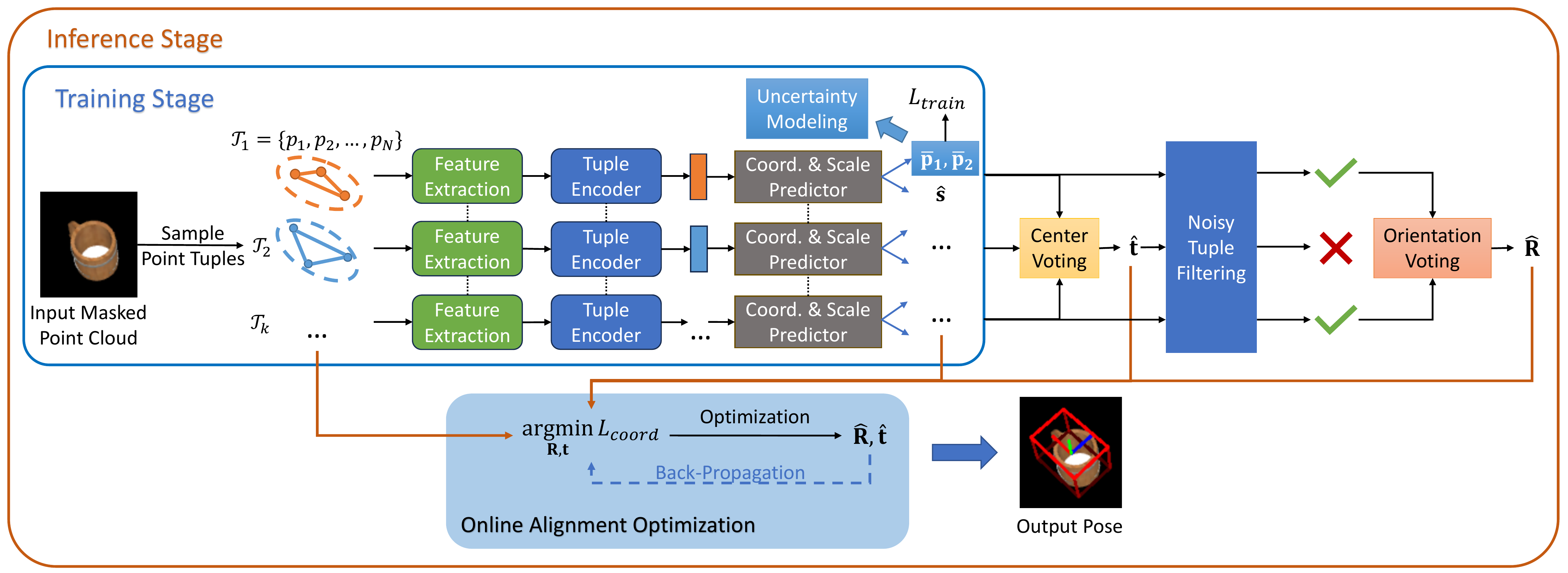}
    \caption{\textbf{Pipeline Overview.} Our pipeline commences with a masked point cloud input, derived from an off-the-shelf instance segmentation model. Subsequently, point tuples are randomly sampled from the object. Features for each tuple are extracted and fed into a tuple encoder to obtain the tuple embedding. Following is the prediction of the canonical coordinate and scale of each tuple. During inference, the computed canonical coordinates and scales are utilized to vote for the object's center. To mitigate the influence of erroneous tuple samples, we introduce a noisy pair filtering module, enabling the orientation vote to be cast exclusively by reliable point tuples. Finally, an online alignment optimization step is employed to further refine the predicted rotation and translation, enhancing the accuracy of our model's output.}
    \label{fig:pipeline}
\end{figure*}
The whole pipeline is given in Figure~\ref{fig:pipeline}. Our network processes a segmented RGB-D backprojected point cloud as input, outputting a 9D pose prediction (3 dimensions for translation, 3 for rotation, and 3 for scale) for the query object. The required input segmentation can be obtained from any off-the-shelf instance segmentation model. While it is feasible to employ state-of-the-art instance segmentation models trained on real-world data, we explore in Section~\ref{sec:maskpred} the possibility of accurately predicting both category-level and instance-level masks without relying on real-world training data. This approach is pivotal in establishing our method as a fully integrated sim-to-real pose estimation framework, eliminating the dependency on real-world training data.

% Finally, Section~\ref{sec:algo} provides a detailed algorithmic description of our entire pipeline, encapsulating the aforementioned components and elucidating the cohesive functioning of our approach.

% Together, these elements constitute a significant advancement in the field of object pose estimation, offering a robust and efficient methodology that is both theoretically sound and practically applicable.

\subsection{Preliminaries: Problem Setting and CPPF Voting}
\label{sec:pre}
The objective of category-level pose estimation is to determine the rotation $\mathbf{R} \in SO(3)$, translation $\mathbf{t} \in \mathbb{R}^3$, and scale $\mathbf{s} \in \mathbb{R}^3_{>0}$ of a target object from an RGB-D image. Consistent with standard practices, the translation corresponds to the \textit{center} (denoted as $\mathbf{o}$) of the object's bounding box, and the rotation is also called the \textit{orientation} of the target object. The terms translation and center, as well as rotation and orientation, will be used interchangeably in the following context.

In the CPPF methodology, an established model for instance segmentation in images, such as MaskRCNN~\cite{he2017mask} or SAM~\cite{kirillov2023segment}, is initially utilized to delineate the object of interest. Upon obtaining the masked point cloud, denoted as $\mathbf{P}$, CPPF~\cite{you2022cppf} selects $K$ point pairs from $\mathbf{P}$. For every point pair, CPPF derives rotation-invariant point pair features for input, subsequently forecasting several voting proxies that align with the object's ground-truth center, orientation, and scale.

\begin{figure}[ht]
\centering
\begin{subfigure}[b]{.45\linewidth}
  \centering
  \includegraphics[width=.9\linewidth]{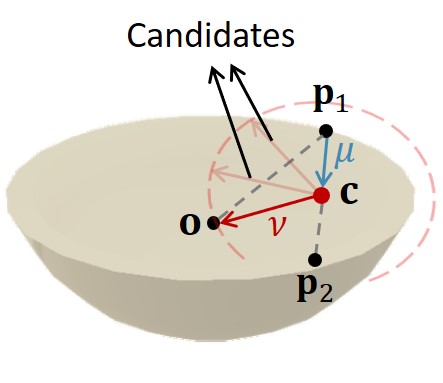}
  \caption{}
  \label{fig:pre1}
\end{subfigure}%
\begin{subfigure}[b]{.45\linewidth}
  \centering
  \includegraphics[width=\linewidth]{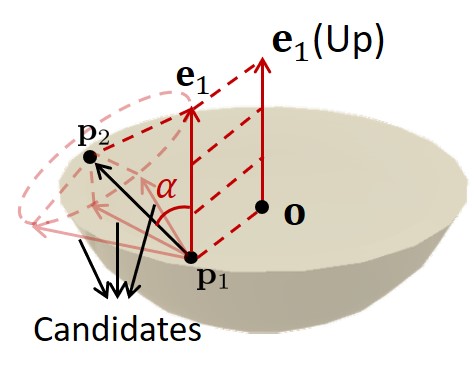}
  \caption{}
  \label{fig:pre2}
\end{subfigure}
\caption{\textbf{(a) Center voting mechanism of CPPF.} Given a point pair $\mathbf{p}_1, \mathbf{p}_2$, CPPF predicts
    $\mu$ and $\nu$, with $\mathbf{c}$ representing the perpendicular foot. \textbf{(b) Orientation voting mechanism of CPPF.} For each point pair, CPPF estimates the angle $\alpha$ between $\mathbf{p}_2 - \mathbf{p}_1$ and $\mathbf{e}_1$. }
\end{figure}

To elucidate further, let the object center be represented as $\mathbf{o}$. For each point pair, $\mathbf{p}_1$ and $\mathbf{p}_2$, CPPF estimates the subsequent two offsets:
\begin{align}
    \mu &= (\mathbf{o} - \mathbf{p}_1)\cdot \mathbf{d}, \label{eq:mu}\\
    \nu &= \|\mathbf{o} - (\mathbf{p}_1 + \mu\mathbf{d}) \|_2, \label{eq:nu}\\
    \text{where\quad}& \mathbf{d} = \frac{\mathbf{p}_2 - \mathbf{p}_1}{\|\mathbf{p}_2 - \mathbf{p}_1\|_2},
\end{align}
as depicted in Figure~\ref{fig:pre1}. It is ensured that the ground-truth center resides on the sphere with its center at $\mathbf{c}$ and radius $\|\mathbf{o}-\mathbf{c}\|$. This voting target generation procedure will henceforth be referred to as:
\begin{align}
    (\mu,\nu) = g(\mathbf{o},\mathbf{p}_1,\mathbf{p}_2). \label{eq:g}
\end{align}

During the inference phase, CPPF uniformly samples point pairs and subsequently enumerates votes around the circle. The predicted center, $\hat{\mathbf{o}}$, can be expressed as a function of the voting targets:
\begin{align}
    \hat{\mathbf{o}} &= f(\hat{\mu},\hat{\nu},\hat{\sigma},\mathbf{p}_1,\mathbf{p}_2)\\
    &= \mathbf{p}_1 + \hat{\mu}\mathbf{d} + \hat{\nu}(\cos(\sigma)\mathbf{d}_\bot+\sin(\sigma)\mathbf{d}\times\mathbf{d}_\bot)\\
    &\text{where\quad} \mathbf{d} = \frac{\mathbf{p}_2 - \mathbf{p}_1}{\|\mathbf{p}_2 - \mathbf{p}_1\|_2},
\end{align}
where $\mathbf{d}_\bot$ represents an arbitrary unit vector orthogonal to $\mathbf{d}$, and $\hat{\sigma}$ is the sampled radial offset in the candidate circle, as illustrated by the red dashed circle in Figure~\ref{fig:pre1}. The location with the predominant vote is designated as the final center prediction.

Regarding orientation voting, CPPF casts votes for the object's upward and rightward orientation basis. The remaining basis is derived using a cross product. With the upward orientation symbolized as $\mathbf{e}_1$ and the rightward orientation as $\mathbf{e}_2$, CPPF predicts the subsequent two relative angles:
\begin{align}
    \alpha &= \mathbf{e}_1\cdot\frac{\mathbf{p}_2 - \mathbf{p}_1}{\|\mathbf{p}_2 - \mathbf{p}_1\|_2} \label{eq:alpha}\\
    \beta &= \mathbf{e}_2\cdot\frac{\mathbf{p}_2 - \mathbf{p}_1}{\|\mathbf{p}_2 - \mathbf{p}_1\|_2},
~\end{align}
as illustrated in Figure~\ref{fig:pre2}. Analogous to the center voting mechanism, CPPF enumerates votes around the red dashed cone, and the orientation amassing the highest number of votes is selected as the final prediction.

For scale prediction, CPPF estimates the scale, denoted as $\mathbf{s}$, for each point pair and subsequently computes their average during the inference phase.

\subsection{A Probabilistic Uncertainty Model of Point Pair Voting}
\label{sec:probabilistic}
CPPF formulates rotation-invariant voting proxies, casting votes for a prospective center, orientation, and scale for each point pair. Without compromising generality, during center voting and given the point pair $\mathbf{p}_1, \mathbf{p}_2$, the ground-truth object center, denoted as $\mathbf{o}$, is unequivocally determined by parameters $\mu,\nu,\sigma$ through the function $f(\mu,\nu,\sigma,\mathbf{p}_1,\mathbf{p}_2)$. Here, $\mu$ and $\nu$ are the proxies delineated in Equations~\ref{eq:mu} and~\ref{eq:nu}, respectively, while $\sigma$ represents the ground-truth radial angle offset on the sphere centered at $\mathbf{c}$ with a radius of $|\mathbf{o}-\mathbf{c}|$. Note that CPPF does not directly predict $\sigma$ but samples it uniformly during the inference phase.

Examining this from a probabilistic perspective, the voting procedure endeavors to optimize the probability of the predicted center, $\hat{\mathbf{o}}$:

\begin{align}
    p(\hat{\mathbf{o}}) = \int &\delta(\hat{\mathbf{o}} - f(\mu,\nu,\sigma,\mathbf{p}_1,\mathbf{p}_2))p(\mu,\nu,\sigma|\mathbf{p}_1, \mathbf{p}_2)\\
    &\cdot p(\mathbf{p}_1, \mathbf{p}_2)d\mathbf{p}_1 d\mathbf{p}_2. \label{eq:prob}
\end{align}
Here, $\delta$ denotes the Dirac delta function. Introducing a naive Bayes prior, under the assumption that $\mu,\nu$, and $\sigma$ are conditionally independent, Equation~\ref{eq:prob} can be reformulated as:
\begin{align}
    p(\hat{\mathbf{o}}) = \int &\underbrace{\delta(\hat{\mathbf{o}} -f(\mu,\nu,\sigma,\mathbf{p}_1,\mathbf{p}_2))}_\textrm{\circled{4}}\underbrace{p(\mu,\nu|\mathbf{p}_1, \mathbf{p}_2))}_\textrm{\circled{2}}\underbrace{p(\sigma|\mathbf{p}_1, \mathbf{p}_2)}_\textrm{\circled{3}}\\
    & \cdot \underbrace{p(\mathbf{p}_1, \mathbf{p}_2)d\mathbf{p}_1 d\mathbf{p}_2}_\textrm{\circled{1}}.
\end{align}

This equation elucidates the operations of CPPF's voting mechanism: \circled{1} for each uniformly sampled point pair $\mathbf{p}_1, \mathbf{p}_2$, \circled{2} the system predicts $\mu,\nu$ utilizing a neural network, and \circled{3} uniformly samples $\sigma$, operating under the assumption that $\sigma$ is independent of $\mathbf{p}_1, \mathbf{p}_2$ (i.e., $p(\sigma|\mathbf{p}_1,\mathbf{p}_2) = p(\sigma)$). Subsequently, \circled{4} the definitive center is determined by $f(\mu,\nu,\sigma,\mathbf{p}_1,\mathbf{p}_2)$, as illustrated in Figure~\ref{fig:pre1}.

\begin{figure}[ht]
    \centering
    \includegraphics[width=\linewidth]{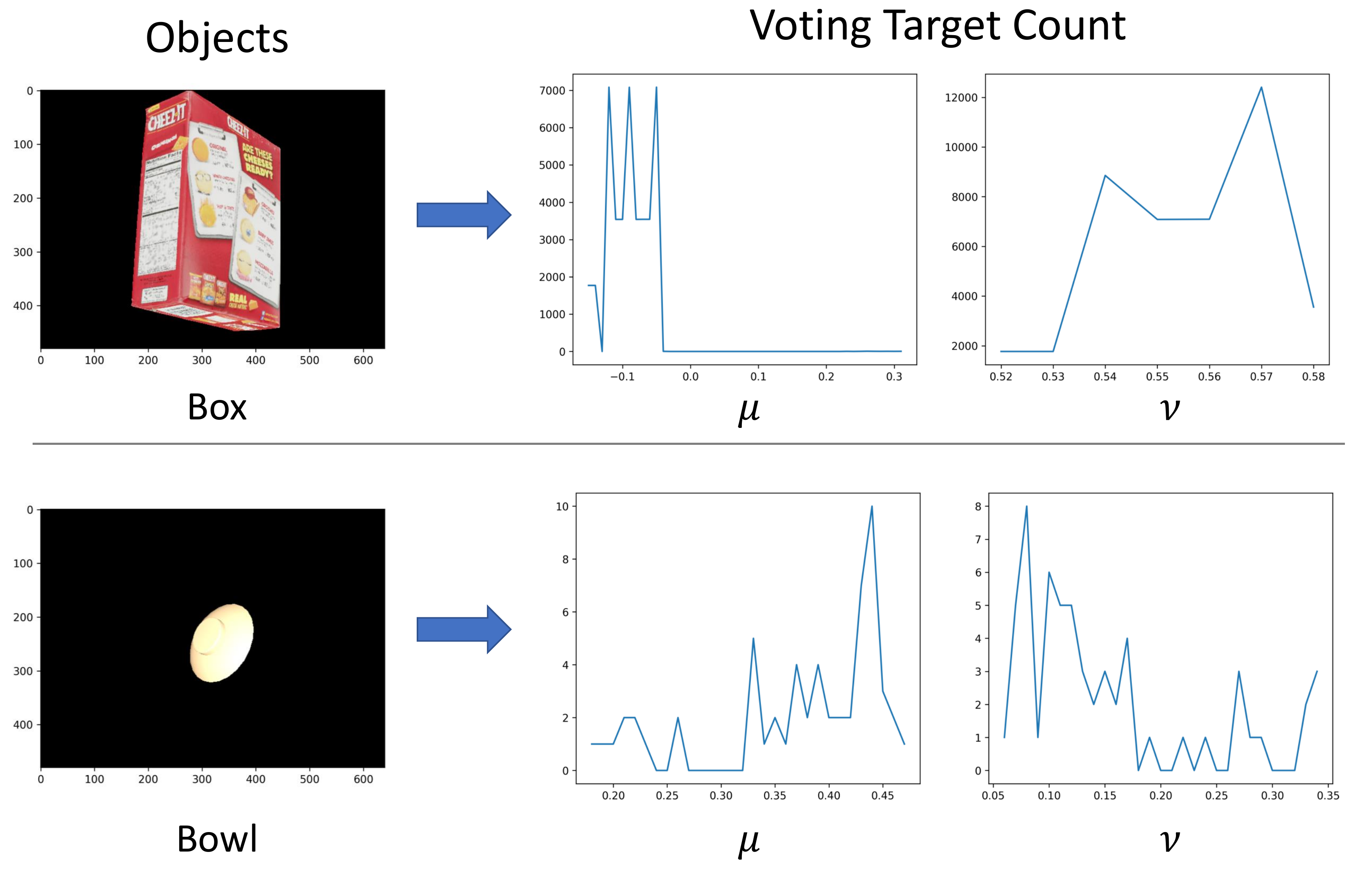}
    \caption{\textbf{This figure illustrates the phenomenon of voting collision for two specific objects: a snack box and a bowl.} For each object, we compile a collection of input pairs characterized by similar input features. This similarity is determined by hashing the features, grouping all point pairs that fall into the same bin. Subsequently, we discretize and count the occurrences of various voting targets $\mu, \nu$ associated with these similar point pairs. The ideal scenario would be for all output voting targets to align, manifesting as a singular, prominent peak in the graph. Contrary to this ideal, the graph reveals multiple peaks, indicating that point pairs with similar input features often lead to significantly disparate voting targets. This variability underscores the difficulty in predicting a consistent deterministic relationship between input features and output voting targets, highlighting the inherent challenges posed by voting collision.}
    \label{fig:collision}
\end{figure}

% \subsection{Vote Uncertainty Modeling}
% \label{sec:uncertainty}
In CPPF, $\mu$ and $\nu$ are predicted by a network that takes as input rotation-invariant features derived from points $\mathbf{p}_1$ and $\mathbf{p}_2$. These features encompass the distance between two point locations and the dot product of their normals. Although these features are rotation invariant, they sacrifice information related to global coordinates, leading to a challenge called ``voting collision.''

Voting collision occurs when distinct pairs of points exhibit similar features but are associated with different voting targets (i.e., $\mu, \nu$). This similarity can confuse the network, making it challenging to assign the correct, distinct voting targets based on the same input feature. This phenomenon is depicted in Figure~\ref{fig:collision}, which illustrates the distribution of voting targets for sets of point pairs with similar features. For commonly encountered objects, such as snack boxes and bowls, the voting collision ratio is significantly high, detrimentally affecting the voting mechanism. Therefore, it is advantageous to represent the output voting targets as a distribution rather than deterministic scalars. We address this challenge by proposing a canonical space probablistic decomposition of $p(\mu,\nu|\mathbf{p}_1,\mathbf{p}_2)$, aiming to explicitly account for this inherent uncertainty:
\begin{align}
    p(\mu,\nu|\mathbf{p}_1,\mathbf{p}_2) &= \delta((\mu,\nu)-g(\mathbf{o},\mathbf{p}_1,\mathbf{p}_2))\\ &= \delta((\mu,\nu)-g(\bar{\mathbf{o}},\bar{\mathbf{p}}_1,\bar{\mathbf{p}}_2))p(\bar{\mathbf{p}}_1,\bar{\mathbf{p}}_2|\mathbf{p}_1,\mathbf{p}_2), 
    \label{eq:nocs}
\end{align}
where $g(\cdot)$ is defined in Equation~\ref{eq:g}, and $\bar{\mathbf{p}}_1,\bar{\mathbf{p}}_2$ are the canonical coordinates of $\mathbf{p}_1,\mathbf{p}_2$. Suppose the ground-truth rotation and translation are $\mathbf{R}$ and $\mathbf{t}$, then $\mathbf{R}\cdot\bar{\mathbf{p}}_1+\mathbf{t}=\mathbf{p}_1$ and $\mathbf{R}\cdot\bar{\mathbf{p}}_2+\mathbf{t}=\mathbf{p}_2$ hold true. We also leverage the fact that $g(\mathbf{o},\mathbf{p}_1,\mathbf{p}_2)) = g(\bar{\mathbf{o}},\bar{\mathbf{p}}_1,\bar{\mathbf{p}}_2))$ because of rotation invariance in the voting targets. $\bar{\mathbf{o}}=(0, 0, 0)$ is the object center in the canonical frame.

The probability distribution $p(\bar{\mathbf{p}}_1,\bar{\mathbf{p}}_2|\mathbf{p}_1,\mathbf{p}_2)$ models the correspondence of input point pairs $\mathbf{p}_1,\mathbf{p}_2$ to canonical point pairs. Ideally, this distribution would be deterministic, represented by a Dirac-$\delta$ distribution. However, due to vote collision, given the input point pairs $\mathbf{p}_1$ and $\mathbf{p}_2$, it is challenging to identify the corresponding canonical point pair based solely on ambiguous point pair features, unless absolute coordinates are provided, which compromises robustness and generalizability. Therefore, in practice we interpret $p(\bar{\mathbf{p}}_1,\bar{\mathbf{p}}_2|\mathbf{p}_1,\mathbf{p}_2)$ as a multinomial distribution over discretized 3D grids within the canonical space. Upon determining the canonical coordinates $\bar{\mathbf{p}}_1,\bar{\mathbf{p}}_2$, we can derive $\mu,\nu$ using $g(\bar{\mathbf{o}},\bar{\mathbf{p}}_1,\bar{\mathbf{p}}_2))$. 
% This is feasible since the ground-truth center $\bar{\mathbf{o}}$ of the canonical mesh is consistently $(0, 0, 0)$, and both $\mu$ and $\nu$ remain invariant to $SE(3)$ transformations.

In the context of orientation voting, we decompose $p(\alpha,\theta|\mathbf{p}_1, \mathbf{p}_2)$ into $ \delta((\alpha,\theta) - g(\mathbf{e}_1,\bar{\mathbf{p}}_1,\bar{\mathbf{p}}_2))p(\bar{\mathbf{p}}_1,\bar{\mathbf{p}}_2|\mathbf{p}_1,\mathbf{p}_2)$. Here, $\alpha$ is the target as defined in Equation~\ref{eq:alpha}, while $\theta$ represents the residual radial degree of freedom, as illustrated in Figure~\ref{fig:pre2}. As our experimental results will demonstrate, this uncertainty modeling significantly enhances performance, rendering the voting procedure more precise.

Regarding scale voting, given that a singular global scaling applies to all point pairs, we adopt the CPPF methodology, predicting the scale $\hat{\mathbf{s}}$ for each point pair and subsequently computing their average during the inference phase.

In our model, we output $\bar{\mathbf{p}}_1,\bar{\mathbf{p}}_2$ and $\hat{\mathbf{s}}$ with a coordinate and scale predictor network respectively. The input to the network will be discussed in Section~\ref{sec:npoint}.

\subsection{$N$-Point Tuple Feature Extraction}
\label{sec:npoint}
CPPF's input features and output voting targets exhibit invariance to arbitrary translations and rotations, making its voting mechanism readily adaptable to real-world contexts. Nevertheless, we observed that the point pair features delineated in CPPF lack enough context required to differentiate between various voting outputs. To enhance the context information and mitigate voting collision, we have expanded the rudimentary point pair features to encompass $N$-point tuple features.

To elaborate, we commence by sampling $K$ $N$-point tuples from the object. We denote each tuple as $\mathcal{T}=\{\mathbf{p}_1,\mathbf{p}_2,\cdots,\mathbf{p}_N\}$. We also estimate each point's normal by local neighborhoods, denoted as $\mathbf{n}_i$. For each $N$-point tuple, we concatenate the following three features as input to our network:
\begin{align}
    \mathcal{F}_1 &= \mathbf{concat}(\{\mathbf{p}_i - \mathbf{p}_j|(i,j)\in \sigma^2(N)\}) \\
    \mathcal{F}_2 &= \mathbf{concat}(\{\max(\mathbf{n}_i \cdot \mathbf{n}_j, -\mathbf{n}_i \cdot \mathbf{n}_j)|(i,j)\in \sigma^2(N)\}) \\
    \mathcal{F}_3 &= \mathbf{concat}(\psi_i|i = 1,2,3,\cdots,N\}),
\end{align}
where $\sigma^2(N)$ denotes all combinations of order 2 from $N$ (commonly referred to as "$N$ choose 2"), $\mathbf{concat}$ signifies concatenation. $\psi_i$ is our proposed additional geometric/visual features for each point. It is chosen from SHOT~\cite{salti2014shot} descriptors and DINOv2~\cite{oquab2023dinov2} features, which will be discussed in Section~\ref{sec:ensemble}.

Feature $\mathcal{F}_1$ encompasses \textit{relative} coordinates, $\mathcal{F}_2$ integrates \textit{relative} normal angles, and $\mathcal{F}_3$ captures the additional geometric/visual clues for each point. In practice, employ random rotation and translation augmentation to the input models to ensure that the features are rotation and translation invariant.

\begin{figure}[ht]
    \centering
    \includegraphics[width=0.9\linewidth]{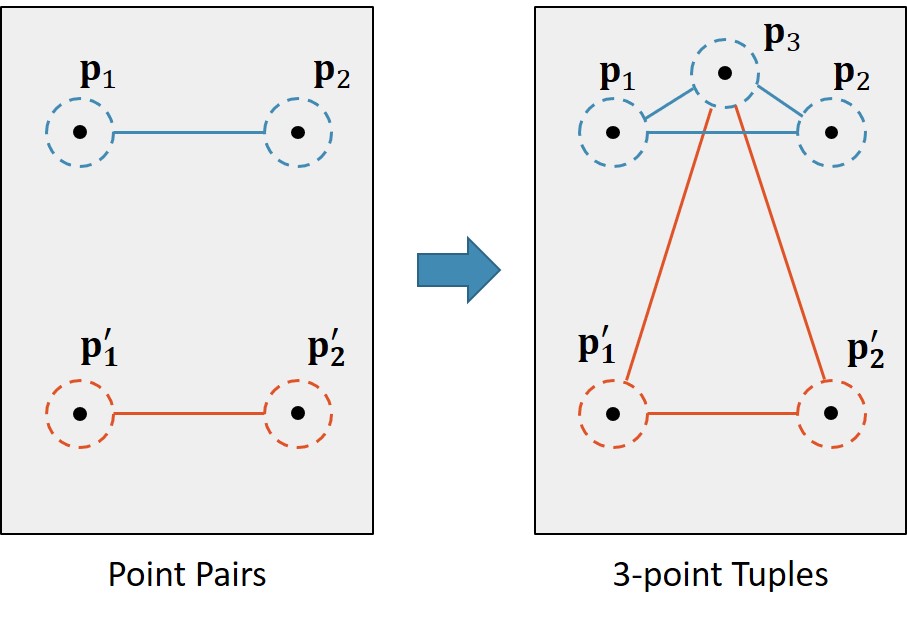}
    \caption{\textbf{Contrast between Point Pairs and $N$-point Tuples.} The dashed circle represents the local context surrounding each point. \textbf{Left:} The two point pairs, $\mathbf{p}_1,\mathbf{p}_2$ and $\mathbf{p}'_1,\mathbf{p}'_2$, possess identical relative coordinates and local contexts, rendering them indiscernible. \textbf{Right:} Conversely, by introducing an extra point, $\mathbf{p}_3$, and creating 3-point tuples using the initial point pairs, the resulting 3-point tuples can now be differentiated based on their relative coordinates.}
    \label{fig:ntuple}
\end{figure}

To elucidate why $N$-point tuples offer more comprehensive information compared to basic point pairs, we present a succinct example in Figure~\ref{fig:ntuple}. On the left side of the figure, two point pairs appear identical when assessed based on their relative coordinates. Distinguishing between these two pairs becomes unfeasible without the inclusion of additional points. However, by introducing another point, say $\mathbf{p}_3$, and constructing 3-point tuples for each original point pair, differentiation becomes possible by calculating the relative coordinates in relation to this additional point.

During training, we first sample point tuples $\mathcal{T}_1, \mathcal{T}_2, ..., \mathcal{T}_k$ from the input RGB-D iamge, then for each tuple, we extract its feature and feed them into our tuple encoder. The tuple encoder network takes the concatenated $\mathcal{F}_1,\mathcal{F}_2,\mathcal{F}_3$ and output the tuple embedding. The tuple embedding is then passed through a coordinate and scale predictor to give the canonical coordinates $\bar{\mathbf{p}}_1,\bar{\mathbf{p}}_2$ for the first two points in the tuple. It also outputs the predicted scale $\hat{\mathbf{s}}$, as shown in Figure~\ref{fig:pipeline}. \textbf{Note that we only use tuples for input feature augmentation but still use point pairs (i.e., the first two points in each tuple) to vote for the poses.}

The training loss consists of two parts, the mean squared error (MSE) of predicted canonical coordinates and the MSE of predicted scales:

\begin{align}
\label{eq:align}
    L_{train} &= L_{coord} + L_{scale}\\
    &= \sum_{\mathcal{T}}\sum_{i=1,2}\|\mathbf{R}^{-1}(\mathbf{p}_i - \mathbf{t}) -\sum_{\mathcal{T}}\bar{\mathbf{p}}_i\|_2 + \|\mathbf{s} -\hat{\mathbf{s}} \|_2,
\end{align}

where $\mathbf{R},\mathbf{t},\mathbf{s}$ are the ground-truth rotation, translation and scale respectively. And the final loss is summed over all tuples.

\subsection{Noisy Pair Filtering}
\label{sec:iterative}
In practical applications, object segmentation often falls short of perfection, potentially including background noise or occlusions that require filtering. Such imperfections significantly undermine the efficacy of orientation voting. To mitigate this, we incorporate a noise filtering module, utilizing the current object center estimate for reference.

To identify and exclude unreliable point pairs, we compute the error of $\mu,\nu$ for each point pair based on the current estimate:
\begin{align}
    \epsilon = \|g(\hat{\mathbf{o}},\mathbf{p}_1,\mathbf{p}_2)) -g(\bar{\mathbf{o}},\bar{\mathbf{p}}_1,\bar{\mathbf{p}}_2))\|_2.
\end{align}
Here, $\hat{\mathbf{o}}$ represents the majority's predicted center, while $\bar{\mathbf{o}}=(0, 0, 0)$. For the sampled point pairs {$\mathbf{p}_1^{(k)}, \mathbf{p}_2^{(k)}$}, where $k$ signifies the sample index, we compute all corresponding errors {$\epsilon^{(k)}$}. Point pairs with errors in the top $\tau\times 100\%$ are subsequently discarded. Although this consistency verification can be executed iteratively by updating $\hat{\mathbf{o}}$, empirical observations indicate that one iteration suffices to eliminate the majority of noise. Adopting this approach significantly bolsters the precision of orientation voting, ensuring a more robust voting mechanism.

\subsubsection{Importance Sample Re-weighting}

While the original point pairs are uniformly sampled from the point cloud,  the noisy pair filtering process can disrupt this balance. As a result, certain points may be assigned reduced weights, especially if a significant number of point pairs linked to them are filtered out. In the subsequent phase, operating under the premise that each point should have an equal influence on the final voting outcome, we adjust the weight of each point inversely based on the number of point pair samples connected to it. While one could opt to re-sample the point pairs from the filtered point clouds, this approach incurs additional computational costs, as the newly sampled point pair input features would need to be reprocessed through our network to determine the output voting targets.

To clarify, when we discuss re-weighting a point, we refer to the process wherein all votes produced by point pairs that include this specific point are adjusted by the weights during the majority tally. This process is analogous to employing a weighted probability for point pair samples, represented as $p(\mathbf{p}_1, \mathbf{p}_2)w(\mathbf{p}_1)w(\mathbf{p}_2)$, where $w(\mathbf{p}_1)$ and $w(\mathbf{p}_2)$ denote the re-weighting factors:
\begin{align}
    w(\mathbf{p}_1) &= \frac{1}{|\{\mathbf{p}_1, \mathbf{p}_2'|\mathbf{p}_2'\in\mathbf{P}\}|+\eta} \\
    w(\mathbf{p}_2) &= \frac{1}{|\{\mathbf{p}_1', \mathbf{p}_2|\mathbf{p}_1'\in\mathbf{P}\}|+\eta}.
\end{align}

\begin{figure}[ht]
    \centering
    \includegraphics[width=\linewidth]{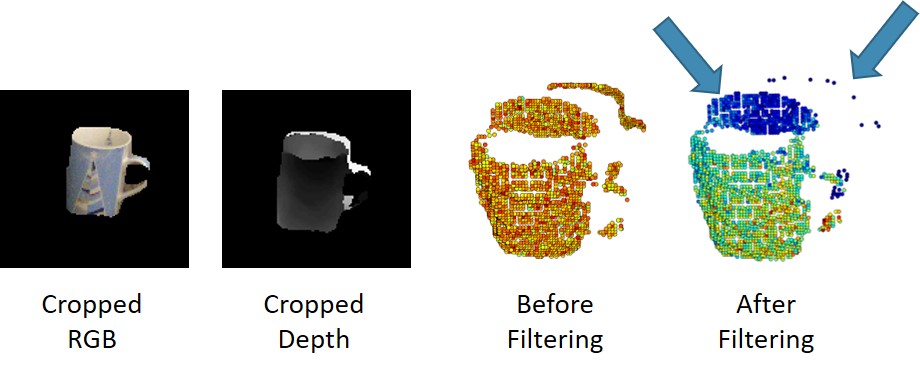}
    \caption{
\textbf{Illustration of Noisy Tuple Filtering and Importance Re-weighting.} Given the detection outcome, background noise may be present, attributable to depth sensor inaccuracies or suboptimal segmentation. On the right, the average sample counts for each point, both pre and post noise filtering, are depicted: red signifies high counts, and blue denotes low. It is evident that the filtering algorithm effectively mitigates the majority of noise but concurrently introduces sampling imbalances on the mug. To maintain uniform sampling, we adjust the weight of each sample inversely based on their counts, obviating the need for re-sampling.}
    \label{fig:filter}
\end{figure}

The notation $|\cdot|$ represents the cardinality of a set, while $\eta$ serves as a margin value, introduced to stabilize the denominator in instances where it encounters small numbers. An illustrative example is provided in Figure~\ref{fig:filter}: prior to noise filtering, all point pairs are assigned relatively uniform weights, and some background noise is evident. Following the noise filtering process, a significant portion of this noise is eradicated. However, certain regions of the mug receive fewer samples compared to others. This distribution is not ideal, as ideally, each point should have an approximately equal influence on the final pose determination.

\subsection{Online Alignment Optimization}
\label{sec:opt}
Our approach predicts voting targets such as $\mu, \nu, \alpha, \beta$, rather than directly estimating $\mathbf{R}$ and $\mathbf{t}$. This strategy ensures that each vote remains independent, significantly enhancing the method's robustness to noise. However, this approach might compromise performance due to the network's limited global perspective on all votes, potentially leading to drift and inaccuracies in the final output.

To address this and further refine the predicted rotation and translation, we introduce an online alignment optimization module. This module utilizes the coordinate loss described in Equation~\ref{eq:align}, but rather than optimizing $\bar{\mathbf{p}}_i$, it directly optimizes the rotation and translation parameters during the online inference phase:
\begin{align}
\hat{\mathbf{R}},\hat{\mathbf{t}} &= \argmin_{\mathbf{R},\mathbf{t}} L_{coord} \\
&= \sum_{\mathcal{T}}\sum_{i=1,2} \|\mathbf{R}^{-1}(\mathbf{p}_i - \mathbf{t}) - \bar{\mathbf{p}}_i\|_2
\end{align}

Given the differentiability of the loss function, we employ the Adam optimizer for the optimization of rotation and translation parameters.

\subsection{Tuple Feature Ensemble}
\label{sec:ensemble}
In Section~\ref{sec:npoint}, we explored the extraction of features from tuples and the computation of an additional $\psi_i$ for each point within the tuple. We evaluated two leading-edge descriptors from the domains of 3D geometry and 2D vision to encapsulate the information surrounding a point. For capturing 3D geometric information, we employ SHOT~\cite{salti2014shot}, a local 3D descriptor known for its effective blend of Signatures and Histograms. This descriptor excels in balancing descriptive capability with robustness. However, it does not encapsulate the semantic information present in the RGB images. To help enriching the feature, we incorporate 2D visual semantics using DINOv2~\cite{oquab2023dinov2}, a self-supervised learning model adept at extracting per-pixel feature descriptors from RGB images.

Our experimental findings suggest that merely concatenating these two descriptors does not achieve optimal performance. This shortfall is attributed to scenarios where an object's pose is determined solely by its geometry, irrespective of its texture, such as a bottle, and a can. On the other side, colors can distinguishes features unidentifiable through geometry alone, like the lid and keyboard of a laptop. Addressing this, we propose to train two separate models, each dedicated to encoding either geometric or visual cues. This enables us to leverage a more powerful ensemble model during inference. Specifically, we conduct a forward pass on both models and compare their optimized $L_{coord}$ losses. The model yielding the lower $L_{coord}$ loss is considered to provide the final prediction. This strategy has significantly enhanced the performance of our model.

\subsection{Instance Mask Prediction}
\label{sec:maskpred}
A salient advantage of our method lies in its reliance solely on synthetic images rendered with CAD models during the training phase. However, our approach does necessitate instance segmentation to sample point tuples on objects, and the instance segmentation module may require additional real-world RGB training data.

Fortunately, recent advancements in zero-shot segmentation methods, such as SAM~\cite{kirillov2023segment} and Grounded DINO~\cite{liu2023grounding}, have demonstrated promising performance in real-world scenarios. In this section, we elucidate how to obtain reliable instance segmentation without the need for additional real-world training data, thereby transforming our method into a comprehensive sim-to-real pipeline.

For category-level detection, we employ Grounded FastSAM~\cite{liu2023grounding,zhao2023fast}, utilizing the text description for each category as the prompt (e.g., bowl, mug, bottle, etc.). The scenario becomes more intricate for instance-level detection, where both the CAD model and text description of the target object are known a priori. In our experiments on the YCB-Video dataset, we observed that the text description of each instance does not consistently facilitate the retrieval of the target instance. Consequently, we propose a two-stage one-shot detection algorithm: 1) We first recall all potential target objects by employing the union of the text prompts from all instances, with Grounded FastSAM~\cite{liu2023grounding,zhao2023fast}. 2) Then, we fine-tune the CLIP model~\cite{radford2021learning} using our synthetically rendered RGB images to classify each crop into distinct instances.

In the interest of a fair comparison, our experimental results include both the outcomes obtained with a commonly used pre-trained instance detector (aligned with previous methods) and those achieved with zero-shot/one-shot instance segmentation methods. This dual reporting underscores the flexibility and robustness of our approach, highlighting its potential for broader application in object pose estimation tasks.

\section{DiversePose 300: A New Dataset with Diverse Poses and Backgrounds}
Current state-of-the-art pose estimation datasets often exhibit a lack of diversity in either backgrounds or poses. To address this gap and offer a more comprehensive evaluation framework, we introduce the DiversePose 300 dataset. This dataset encompasses three prevalent object categories: bowls, bottles, and mugs. Each category is depicted through 100 annotated frames, culminating in a total of 300 unique frames. This approach stands in contrast to the NOCS REAL275 dataset, characterized by its reliance on only 6 continuous videos for evaluation purposes.

The DiversePose 300 dataset has been carefully crafted to promote greater diversity in pose estimation by allowing objects to rotate freely, rather than being constrained to an upright position on a table. The dataset is organized into three distinct scenarios, each designed to represent a varying level of difficulty:

\begin{itemize}
    \item \textbf{Easy:} A single object is placed on the desktop with a relatively clean background.
    \item \textbf{Medium:} A single object is held by hand, with a relatively clean background.
    \item \textbf{Hard:} Multiple objects are scattered on the material box or desktop, and the background is relatively complex.
\end{itemize}

\begin{figure*}[ht]
    \centering
    \includegraphics[width=0.9\linewidth]{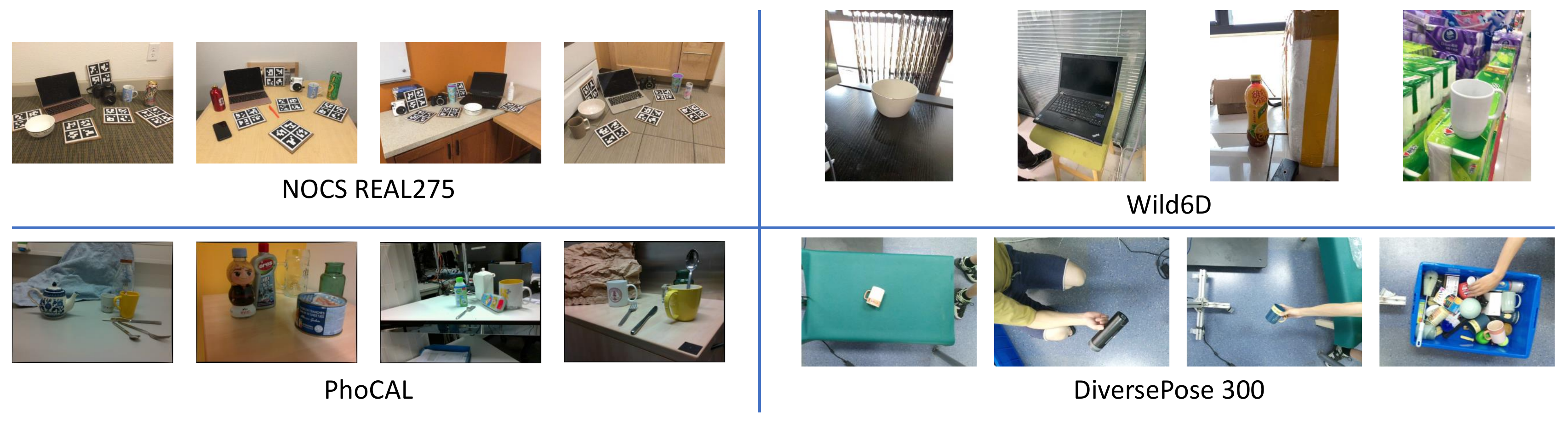}
    \caption{\textbf{A qualitative comparison between various datasets.} NOCS REAL275 exhibits a bias towards similar backgrounds and predominantly upright poses. Wild6D and PhoCAL, while offering more diverse backgrounds, also display a bias towards upright poses. In contrast, our dataset, DiversePose 300, is specifically designed to encompass a broad spectrum of pose variations against varied backgrounds and levels of difficulty, addressing the limitations observed in the aforementioned datasets.}
    \label{fig:dataset}
\end{figure*}

We annotated the pose of each object using an open-source annotator for point clouds~\cite{Sager_2022}. This tool offers a user-friendly interface that facilitates interactive inspection and annotation of oriented bounding boxes in 3D.

A qualitative comparison, as illustrated in Figure~\ref{fig:dataset}, between the DiversePose 300 dataset and several existing pose estimation datasets underscores substantial variations in pose and background diversity. Despite its relatively compact size, DiversePose 300 poses a significant challenge to contemporary state-of-the-art approaches, with performance metrics far from reaching a plateau. Our experimental results demonstrate that our method significantly surpasses existing state-of-the-art techniques across all these unseen datasets.

% The distinctions between DiversePose 300 and NOCS REAL275 can be summarized in three key aspects:

% \begin{itemize}
%     \item \textbf{Annotation Diversity:} Unlike the continuous videos in NOCS REAL275, DiversePose 300's individual unrelated frames result in more diverse pose annotations.
%     \item \textbf{Robustness of Evaluation:} The inclusion of challenging scenarios such as bin clutters and hand manipulation emphasizes the evaluation of algorithm robustness.
%     \item \textbf{Depth Rendering Complexity:} The utilization of the Intel RealSense camera, as opposed to the Structure Sensor used in NOCS REAL275, creates more challenging depth renderings, necessitating that evaluated methods generalize across different data domains.
% \end{itemize}

In conclusion, DiversePose 300 offers a unique and challenging dataset, complement to the comprehensive evaluation of object pose estimation. Its design emphasizes pose and background diversity, setting it apart from existing datasets.

\section{Implementation Details}

In our approach, we opt for $5$-point tuples and set a filtering ratio of $\tau=0.5$, with re-weighting margin $\eta=1.0$. Within our model, each $5$-point tuple is processed independently to predict a voting proxy. We use the SHOT implementation from PCL~\cite{rusu20113d} library and the pretrained DINOv2 ViT-L/14~\cite{oquab2023dinov2} features. For the tuple encoder, we use $128 \times 5$ hidden residual MLP layers, with $256$ neurons in the final layer. We use MLP with 2 hidden layers to predict the canonical coordinates, with an output dimension of $2 \times 3 \times 32$. It represents the distribution of the canonical coordinate for the first two points in the tuple, and the distribution is discretized into 32 bins within $[-1,1]$ across the three $xyz$ axes.
% \begin{itemize}
%     \item \textbf{SHOT Feature Encoder:} This encoder is responsible for compactly embedding each point's SHOT descriptor. It consists of five residual MLP layers, following the architecture proposed by He \textit{et al.}~\cite{he2016deep}.
%     \item \textbf{Tuple Encoder:} Comprising $128 \times 5$ hidden residual MLP layers, the tuple encoder culminates in a final layer with $256$ neurons. Subsequently, either the logit encoder or the scale encoder is concatenated to this structure.
%     \item \textbf{NOCS Encoder:} This encoder features two hidden layers and produces $2 \times 3 \times 32$ outputs, representing the distribution of the canonical coordinate for each point pair. The distribution is discretized into 32 bins across the three $xyz$ axes.
%     \item \textbf{Scale Encoder:} Comprising two hidden layers with $256$ and $128$ neurons, the scale encoder ultimately yields three outputs, predicting the $xyz$ scale for each tuple.
% \end{itemize}

Training is conducted separately for each category or instance, utilizing the Adam optimizer with a learning rate of $1\times 10^{-3}$. The training process spans 100 epochs, with the learning rate being halved every 25 epochs. Each epoch contains 200 samples.

During the training stage, to emulate self-occlusions, we adhere to the method employed by CPPF, leveraging Pyrender with an OpenGL back-end~\cite{matl2018} to synthesize projected point clouds online. Our method utilizes the identical synthetic ShapeNet objects as CPPF~\cite{shapenet2015}. At the inference stage, our network directly accepts the RGB-D image segmentation as input.

We sample point pairs randomly with uniform distribution during the voting stage. In the center voting process, the accumulation 3D grid is defined with a resolution of 0.2 cm, and its range is determined by the tightest axis-aligned bounding box of the input. During the orientation voting process, the orientation grid is set with a resolution of 1 degree, ensuring precision in the pose estimation. For online alignment optimization, we re-parameterize the rotation and translation into $se(3)$ using LieTorch~\cite{teed2021tangent}, and run gradient descent on this manifold. The optimization learning rate is 1e-2, and we run for 100 iterations.

% This architecture and methodology collectively contribute to a robust and efficient model for object pose estimation, demonstrating the potential for practical applications in various computer vision tasks.

\section{Experiments}
% In this section, we evaluate the proposed method on both category-level and instance-level pose estimation settings.
% , we evaluate our method on both NOCS REAL275 and YCB-Video dataset.
\subsection{Results on NOCS REAL275}
\subsubsection{Datasets.} NOCS REAL275~\cite{wang2019normalized} dataset is a common benchmark used to evaluate our method on category-level pose estimation, where only a collection of similar objects in the same category during training time. NOCS REAL275 dataset captures 2750 test frames of 6 real scenes using a Structure Sensor.

\begin{figure}[ht]
    \centering
    \includegraphics[width=\linewidth]{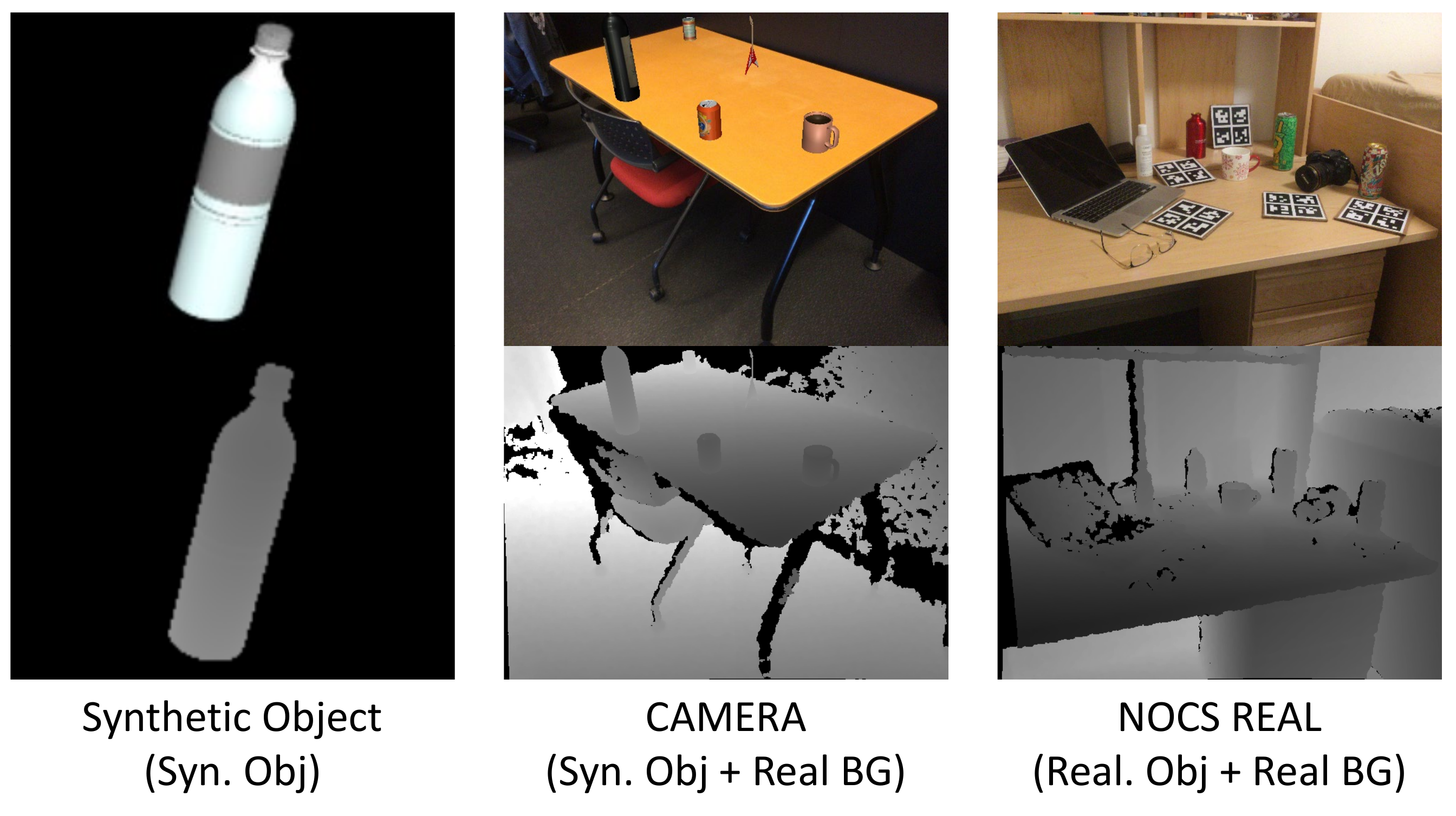}
    \caption{\textbf{RGB-D visualization of different training data modalities.} Our method uses synthetic objects (Syn. Obj) where no background or clutter prior is introduced; CAMERA (Syn. Obj + Real BG) composes images by placing rendered synthetic objects on real backgrounds (i.e., tables); NOCS REAL (Real Obj + Real BG) is a dataset captured with real objects placed in real scenarios.}
    \label{fig:diffnocs}
\end{figure}

\begin{figure*}[ht]
    \centering
    \includegraphics[width=\linewidth]{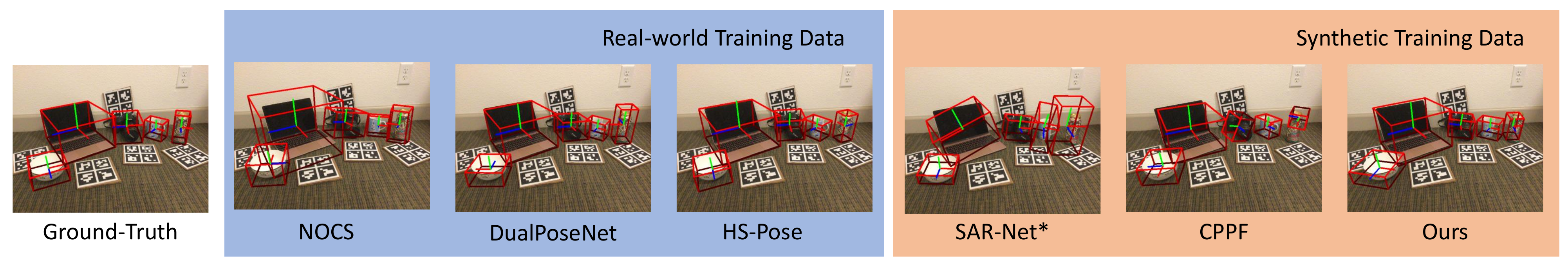}
    \caption{\textbf{Comparison of different methods on NOCS REAL275 dataset.} Though trained with synthetic CAD models, our method is comparable to current state-of-the-arts.}
    \label{fig:resnocs}
\end{figure*}

\subsubsection{Metrics.} 
We adopt the metrics from NOCS~\cite{wang2019normalized} to present results using both intersection over union (IoU) and 6D pose average precision. The IoU is determined by comparing the predicted bounding boxes with the ground-truth, using thresholds of 25\% and 50\%. Meanwhile, the 6D pose average precision is ascertained by gauging the average precision for objects where the error is below $m$ cm for translation and $n^\circ$ for rotation.

It is noteworthy that the original code for 3D box mAP computation provided by NOCS contains errors. Consequently, we, in alignment with CPPF, have opted to utilize the code available in Objectron~\cite{ahmadyan2021objectron}. This flawed code has been inadvertently adopted by numerous subsequent studies. As a result, we have taken the initiative to reproduce the results for 3D box mAP specifically for NOCS, DualPoseNet, SAR-Net and HS-Pose, given that they offer reproducible code.

\subsubsection{Baselines.}
Based on the modality of the training data utilized, we categorize the baselines into 4 distinct types: \begin{itemize}
    \item Methods such as NOCS~\cite{wang2019normalized}, CASS~\cite{chen2020learning}, SPD~\cite{tian2020shape}, FS-Net~\cite{chen2021fs}, CenterSnap~\cite{irshad2022centersnap}, ShAPO~\cite{irshad2022shapo}, DualPoseNet~\cite{Lin_2021_ICCV}, and HS-Pose~\cite{zheng2023hs} utilize both the CAMERA synthetic dataset and the NOCS REAL training set offered by NOCS.  The training data consists of synthetic objects (Syn. Obj) + real objects and backgrounds (Real Obj \& BG).
    \item RePoNet~\cite{ze2022category}, UDA-COPE~\cite{lee2022uda} and TTA-COPE~\cite{lee2023tta} use both synthetic and real training data, but they use semi-supervised learning and do not require ground-truth labels from the training data.
    \item SAR-Net~\cite{lin2022sar} leverages the CAMERA synthetic dataset provided by NOCS, wherein synthetic CAD models (Syn. Obj) are randomly placed onto real background (Real BG) tables, as shown in the middle of Figure~\ref{fig:diffnocs}. SAR-Net is heavily dependent on this background prior to filter out noise from the instance mask. 
    \item Chen \textit{et al.}~\cite{chen2020category}, Gao \textit{et al.}~\cite{gao20206d}, and CPPF~\cite{you2022cppf} exclusively use synthetic objects (Syn. Obj), devoid of any background simulation. Our approach aligns with CPPF in using synthetic objects only. We also train a background-free version of SAR-Net*~\cite{lin2022sar}, where no real-world background is provided in the training data. 
\end{itemize}
The distinctions among the Synthetic Object, CAMERA, and NOCS REAL datasets are depicted in Figure~\ref{fig:diffnocs}. Given that the provided mask/detection prior often includes noise or background elements, the task of sim-to-real transfer becomes inherently challenging.
In order to align with previous methods, we use the mask prior provided by DualPoseNet. CPPF was originally evaluated with the mask provided by NOCS, so we re-evaluate it with DualPoseNet's mask.
% These particular methods train their own masks, which tend to be marginally superior to those furnished by NOCS, potentially rendering their results more favorable.

\subsubsection{Results.}

\begin{table}[ht]
\begin{center}
\resizebox{\linewidth}{!}{
\begin{tabular}{lcccccc}
\toprule
\multirow{3}*{} & \multirow{3}*{Training Data} & \multicolumn{5}{c}{mAP (\%)}\\
\cmidrule(lr){3-7}
~ & ~ & \multirowcell{2}{3D$_{25}$} & \multirowcell{2}{3D$_{50}$} & \multirowcell{2}{5$^\circ$\\ 5 cm} & \multirowcell{2}{10$^\circ$\\ 5 cm} & \multirowcell{2}{15$^\circ$\\ 5 cm} \\
& & & & & & \\
\midrule
NOCS~\cite{wang2019normalized}  & Syn. Obj + Real Obj \& BG & 74.4 & 27.8 & 9.8 & 24.1 & 34.9\\
CASS~\cite{chen2020learning} & Syn. Obj + Real Obj \& BG & - & - & 23.5 & 58.0 & - \\
SPD~\cite{tian2020shape} & Syn. Obj + Real Obj \& BG & - & - & 21.4 & 54.1 & - \\
FS-Net~\cite{chen2021fs} & Syn. Obj + Real Obj \& BG & - & - & 28.2 & 60.8 & - \\
CenterSnap~\cite{irshad2022centersnap} & Syn. Obj + Real Obj \& BG  & - & - & 27.2 & 58.8 & - \\
ShAPO~\cite{irshad2022shapo} & Syn. Obj + Real Obj \& BG  & - & - & 48.8 & 66.8 & - \\
DualPoseNet~\cite{Lin_2021_ICCV} & Syn. Obj + Real Obj \& BG  & 82.3 & 57.3 & 36.1 & 67.8 & 76.3 \\
HS-Pose~\cite{zheng2023hs} & Syn. Obj + Real Obj \& BG  & \textbf{82.6} & \textbf{71.6} & \textbf{56.1}& \textbf{84.1} & \textbf{92.8} \\

\midrule
RePoNet~\cite{ze2022category} & Syn. Obj + Real Obj \& BG (w/o label) & - & - & 33.9 & 63.0 & - \\
UDA-COPE~\cite{lee2022uda} & Syn. Obj + Real Obj \& BG (w/o label) & - & - & 34.8 & 66.0 & - \\
TTA-COPE~\cite{lee2023tta} & Syn. Obj + Real Obj \& BG (w/o label) & - & - & 35.9 & 73.2 & - \\
\midrule
SAR-Net~\cite{lin2022sar} & Syn. Obj + Real BG & 77.9 & 47.5 & 42.3 & 68.3 & 75.0 \\
\midrule
Chen \textit{et al.}~\cite{chen2020category} & Syn. Obj & 16.7 & 0.7 & 0.8 & 3.1 & 7.2 \\
Gao \textit{et al.}~\cite{gao20206d} & Syn. Obj & 70.6 & 23.1 & 5.1 & 14.2 & 21.9 \\
SAR-Net*~\cite{lin2022sar} & Syn. Obj & 64.0 & 34.0 & 4.8 & 21.1 & 30.7 \\
CPPF~\cite{you2022cppf} & Syn. Obj & 78.2 & 25.0 & 18.0 & 45.2 & 51.6  \\
\midrule
Ours  & Syn. Obj & \textbf{82.4} & 55.2 & \textbf{32.3} & 65.9 & 85.2 \\
Ours w/ GroundedSAM Det.  & Syn. Obj & 72.7 & \textbf{58.0} & 26.2 & \textbf{72.0} & \textbf{88.9} \\
\bottomrule
\end{tabular}}
\end{center}
\caption{\textbf{Comparison of various methods on NOCS REAL275.} The best methods that use real training data and synthetic data are bolded respectively.
% Our method trains on synthetic objects, and performs well in real scenarios.
}
\label{tab:nocs}
\end{table}

Quantitative outcomes are presented in Table~\ref{tab:nocs}. It is evident that our approach significantly surpasses CPPF across all metrics, registering improvements of \textbf{4.2}, \textbf{30.2}, \textbf{14.3}, \textbf{20.7} and \textbf{33.6} in terms of 3D$_{50}$, 3D$_{25}$, 5$^\circ$5 cm, 10$^\circ$5 cm, and 15$^\circ$5 cm, respectively. Our method also compares to or even output perform several methods that employ real training data. While HS-Pose exhibits commendable performance on NOCS REAL275, as we will discuss in Section~\ref{sec:diverseexp}, its accuracy considerably diminishes on datasets in the wild. Additionally, we present pose estimation results utilizing GroundedSAM~\cite{liu2023grounding,kirillov2023segment} as our detector, which further enhances performance by delivering marginally improved segmentation. CPPF++ effectively bridges the domain gap between sim-to-real and methods trained on real data. Visual comparisons are showcased in Figure~\ref{fig:resnocs}.

\begin{table}[ht]
\begin{center}
% \resizebox{\linewidth}{!}{
\begin{tabular}{lcccccc}
\toprule
\multirow{3}*{} & \multirow{3}*{} & \multicolumn{5}{c}{mAP (\%)}\\
\cmidrule(lr){3-7}
~ & ~ & \multirowcell{2}{3D$_{25}$} & \multirowcell{2}{3D$_{50}$} & \multirowcell{2}{5$^\circ$\\ 5 cm} & \multirowcell{2}{10$^\circ$\\ 5 cm} & \multirowcell{2}{15$^\circ$\\ 5 cm} \\
& & & & &\\
\midrule

\multirow{6}*{NOCS~\cite{wang2019normalized}} & bottle & 42.1 & 15.1 & 2.7 & 12.9 & 28.7\\ 
 & bowl & \textbf{100.0} & 88.6 & 49.7 & 84.3 & 96.4\\ 
 & camera & 79.4 & 13.1 & 0.0 & 0.1 & 0.6\\ 
 & can & 39.0 & 6.7 & 0.4 & 7.9 & 18.0\\ 
 & laptop & \textbf{92.8} & 0.6 & 6.5 & 41.2 & 68.2\\ 
 & mug & 93.6 & 42.8 & 0.0 & 0.1 & 1.9\\ 

\midrule

\multirow{6}*{DualPoseNet~\cite{Lin_2021_ICCV}} & bottle & \textbf{50.2} & \underline{25.8} & 38.2 & 84.9 & 97.5\\ 
 & bowl & \textbf{100.0} & \underline{97.3} & \underline{78.8} & \underline{97.5} & \underline{99.7}\\ 
 & camera & \textbf{90.0} & 33.3 & 0.0 & 0.5 & 3.7\\ 
 & can & 70.2 & 20.9 & \underline{47.0} & 93.0 & \underline{97.7}\\ 
 & laptop & 84.8 & \textbf{82.5} & 49.2 & \underline{88.7} & \underline{95.4}\\ 
 & mug & \textbf{99.6} & 85.1 & \underline{7.1} & \underline{47.9} & 68.7\\ 

\midrule

\multirow{6}*{HS-Pose~\cite{zheng2023hs}} & bottle & 46.8 & \textbf{37.3} & \textbf{53.3} & \textbf{96.9} & \textbf{99.8}\\ 
 & bowl & 99.7 & \textbf{99.7} & \textbf{92.4} & \textbf{99.6} & \textbf{99.8}\\ 
 & camera & 88.4 & \textbf{52.9} & \textbf{5.9} & \textbf{40.8} & \textbf{70.9}\\ 
 & can & \underline{73.7} & \textbf{70.0} & \textbf{70.6} & \textbf{99.4} & \textbf{99.5}\\ 
 & laptop & \underline{86.3} & \underline{80.9} & \textbf{78.7} & \textbf{92.2} & 92.7\\ 
 & mug & \underline{97.7} & \underline{90.8} & \textbf{24.6} & \textbf{77.6} & \textbf{86.2}\\ 

\midrule

\multirow{6}*{SAR-Net*~\cite{lin2022sar}} & bottle & 28.6 & 6.8 & 11.6 & 42.6 & 61.3\\ 
 & bowl & \underline{99.6} & 23.1 & 11.6 & 60.4 & 81.9\\ 
 & camera & 60.7 & 0.5 & 0.0 & 0.0 & 0.0\\ 
 & can & 31.3 & 11.8 & 9.4 & 39.7 & 62.5\\ 
 & laptop & 82.9 & 26.9 & 0.0 & 0.0 & 0.0\\ 
 & mug & 62.4 & 0.0 & 0.0 & 0.6 & 2.4\\ 

\midrule

\multirow{6}*{CPPF~\cite{you2022cppf}} & bottle & 28.6 & 0.2 & 26.7 & 85.7 & 94.4\\ 
 & bowl & 96.8 & 7.1 & 7.7 & 40.1 & 62.9\\ 
 & camera & 88.3 & 4.2 & 0.0 & 0.0 & 0.0\\ 
 & can & \textbf{73.9} & \underline{34.7} & 45.5 & \underline{95.4} & 97.7\\ 
 & laptop & 84.2 & 24.4 & 27.8 & 47.8 & 49.4\\ 
 & mug & 97.6 & 79.6 & 0.1 & 2.0 & 5.0\\ 

\midrule

\multirow{6}*{Ours} & bottle & \underline{48.1} & 22.4 & \underline{42.6} & \underline{96.9} & \underline{99.4}\\ 
 & bowl & \textbf{100.0} & 81.0 & 48.5 & 75.8 & 93.3\\ 
 & camera & \underline{89.7} & \underline{39.8} & \underline{0.4} & \underline{13.4} & \underline{48.8}\\ 
 & can & 70.8 & 12.1 & 26.7 & 78.3 & 94.1\\ 
 & laptop & 85.9 & 80.9 & \underline{71.8} & 87.3 & \textbf{95.7}\\ 
 & mug & \textbf{99.6} & \textbf{95.3} & 3.8 & 43.8 & \underline{80.1}\\ 
\bottomrule
\end{tabular}
% }
\end{center}
\caption{\textbf{Per-category performance of the proposed method on NOCS REAL275.} The best results are \textbf{bolded} and the second-best results are \underline{underscored}. NOCS, DualPoseNet and HS-Pose are real-world training methods while SAR-Net*, CPPF and ours are pure sim-to-real methods.}
\label{tab:nocs_cat}
\end{table}

\subsubsection{Detailed Analysis and Failure Cases.}

\begin{figure}[ht]
    \centering
    \includegraphics[width=0.8\linewidth]{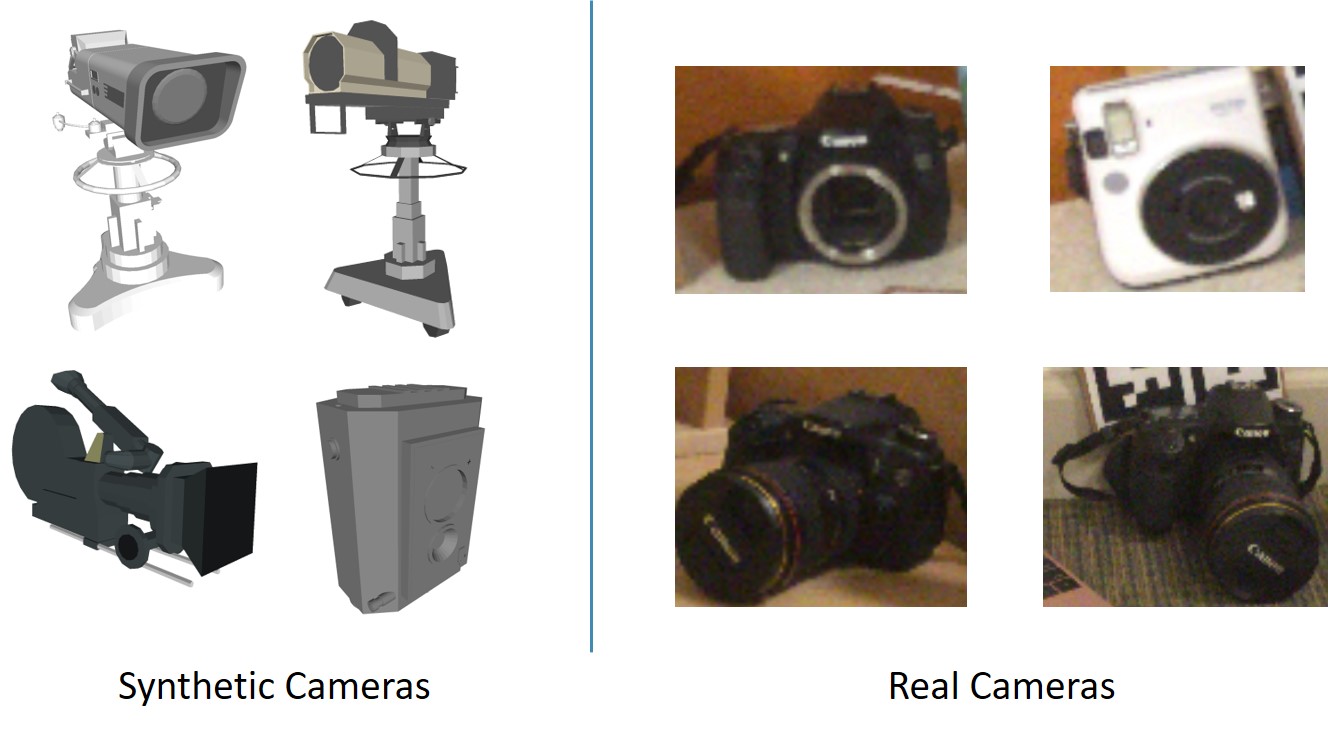}
    \caption{In NOCS REAL275 dataset, there is a distribution shift between synthetic and real cameras, causing a drop of our method on the \textit{camera} category.}
    \label{fig:diff}
\end{figure}

Detailed per-category outcomes are provided in Table~\ref{tab:nocs_cat}. Our methods achieve the best among all sim-to-real methods. The inferior performance associated with the camera category can be attributed to the disparity between synthetic CAD models and actual objects. As illustrated in Figure~\ref{fig:diff}, the CAD models under evaluation and the synthetic CAD models for cameras belong to distinct types. This variance cannot be rectified without employing real-world training data that aligns with the same distribution.

\subsection{Results on Datasets in the Wild}
\label{sec:diverseexp}
Though current state-of-the-art methods get decent results on NOCS REAL275, we are extremely interested in if these methods can generalize to other data distributions. As a result, we compare the performance of our method and state-of-the-art methods on three out-of-distribution datasets: Wild6D~\cite{ze2022category}, PhoCAL~\cite{wang2022phocal} and DiversePose 300 (our proposed). 

\subsubsection{Metrics.}
We use exactly the same metric as NOCS REAL275, while loosen the AP threshold to $20^\circ, 40^\circ, 60^\circ$ for DiversePose 300 since it has a more diverse distribution of pose rotations.

\subsubsection{Baselines.}

We benchmark against a suite of state-of-the-art methods, including NOCS~\cite{wang2019normalized}, DualPoseNet~\cite{Lin_2021_ICCV}, SAR-Net~\cite{lin2022sar}, RePoNet~\cite{ze2022category}, HS-Pose~\cite{zheng2023hs}, and CPPF~\cite{you2022cppf}. It is important to note that RePoNet has not made available the pretrained model for the \textit{can} category, precluding its $can$ evaluation on the PhoCAL dataset.

To facilitate a fair comparison, for DiversePose 300, which features a distinct rotation distribution, we retrained all aforementioned methods with rotation augmentation spanning the entire $SO(3)$ space. This was applicable to all methods except RePoNet, due to the unavailability of its training code. The training dataset consists of synthetically rendered RGB-D frames showcasing randomized orientations and translations of objects.

During the inference phase, all methods employ the ground-truth masks provided by Wild6D and PhoCAL. For DiversePose300, since we only annotated the 9D poses, we use the leading-edge detector, Masked-DINO~\cite{li2023mask}.

\subsubsection{Results.}
\begin{figure*}[ht]
    \centering
    \includegraphics[width=\linewidth]{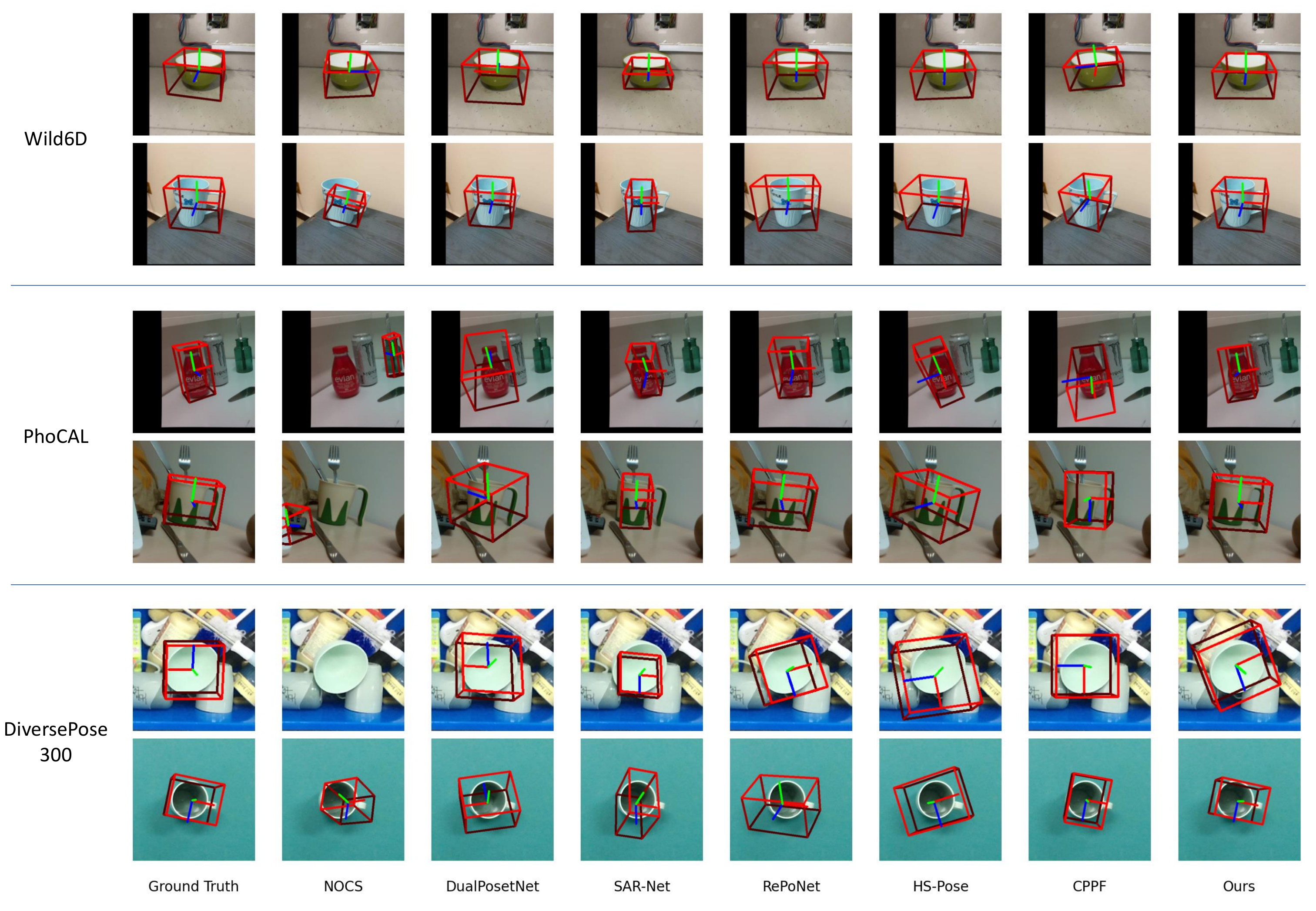}
    \caption{\textbf{Qualitative comparison of different methods on three novel datasets: Wild6D, PhoCAL and DiversePose300.} When transferring to a new data domain, our method shows stronger generalizability than previous methods.}
    \label{fig:wild}
\end{figure*}

Results for Wild6D, PhoCAL and DiversePose 300 are depicted in Table~\ref{tab:wild6d},~\ref{tab:phocal} and ~\ref{tab:diverse} respectively. Our introduced method markedly surpasses preceding methodologies on all three unseen datasets. Baseline methods get struggled with the diverse poses and backgrounds present in the novel scenes. They exhibit a propensity to skew towards the distribution of their training data, whereas our model demonstrates robustness in generalizing to previously unseen data. While CPPF manages to produce some plausible results, it remains inferir compared to our proposed approach. A visual comparison is presented in Figure~\ref{fig:wild}.

% \subsubsection{Failure Cases.}

\begin{table}[!h]
\centering
\begin{tabular}{lcccccc}
\toprule
\multirow{3}*{} & \multirow{3}*{} & \multicolumn{5}{c}{AP (\%)$\uparrow$} \\
\cmidrule(lr){3-7}
~ & ~ &\multirowcell{2}{3D$_{25}$} & \multirowcell{2}{3D$_{50}$} & \multirowcell{2}{5$^\circ$\\ 5 cm} & \multirowcell{2}{10$^\circ$\\ 5 cm} & \multirowcell{2}{15$^\circ$\\ 5 cm} \\\\
\midrule
\multirow{5}*{NOCS~\cite{wang2019normalized}} & bottle & 50.5 & 34.1 & 39.2 & 82.4 & 91.2\\
& bowl & 80.3 & 47.3 & 22.8 & 76.9 & 90.1\\
& camera & 10.4 & 1.7 & 0.0 & 0.0 & 0.0\\
& laptop & 97.1 & 77.1 & 4.3 & 15.0 & 15.9\\
& mug & 16.5 & 0.4 & 0.0 & 0.2 & 1.0\\

\midrule
\multirow{5}*{DualPoseNet~\cite{Lin_2021_ICCV} } & bottle & \textbf{89.5} & 61.7 & 43.2 & 80.1 & 92.1\\
& bowl & \textbf{98.1} & \textbf{89.2} & 67.0 & 89.9 & \underline{93.4}\\
& camera & \textbf{82.5} & \textbf{22.0} & 0.0 & \underline{1.2} & \underline{3.9}\\
& laptop & 98.1 & \underline{97.2} & 12.0 & 14.8 & 15.0\\
& mug & 82.5 & \textbf{63.5} & 0.5 & 3.3 & 5.3\\

\midrule
\multirow{5}*{SAR-Net~\cite{lin2022sar}} & bottle & 86.1 & 28.5 & 5.8 & 44.0 & 82.4\\
& bowl & 93.0 & \textbf{0.5} & 38.1 & 83.2 & 92.4 \\
& camera & 34.8 & 0.2 & 0.0 & 0.0 & 0.2\\
& laptop & 95.7 & 1.1 & 0.0 & 0.0 & 0.0\\
& mug & 78.3 & 0.0 & 0.0 & 0.6 & 2.9 \\

\midrule
\multirow{5}*{RePoNet~\cite{ze2022category}} & bottle & \underline{88.4} & \textbf{76.3} & \textbf{79.2} & \underline{93.9} & \underline{95.7}\\
& bowl & \underline{97.3} &\underline{86.2} & 71.2 & \underline{91.2} & 92.9 \\
& camera & 44.8 & 2.4 & \underline{0.1} & 0.6 & 1.5\\
& laptop & 97.3 & 93.1 & \textbf{14.5} & 18.0 & \textbf{19.4}\\
& mug & 77.9 & \underline{52.0} & \textbf{2.5} & \textbf{15.1} & \textbf{23.0} \\

\midrule
\multirow{5}*{HS-Pose~\cite{zheng2023hs}} & bottle & 87.1 & \underline{70.4} & \underline{54.3} & 89.1 & 94.1 \\
& bowl & 96.5 & 82.3 & \underline{74.0} & 89.4 & 92.0 \\
& camera & 59.8 & 5.4 & 0.0 & 0.3 & 1.1\\
& laptop & \textbf{99.7} & 90.0 & 12.6 & 17.0 & 17.1 \\
& mug & \textbf{83.8} & 42.4 & \underline{2.0} & \underline{12.3} & \underline{19.3} \\

\midrule
\multirow{5}*{CPPF~\cite{you2022cppf}} & bottle & 74.9 & 10.3 & 22.4 & 44.2 & 46.4 \\
& bowl & 91.2 & 14.3 & 11.8 & 35.2 & 49.2 \\
& camera & 53.4 & 0.1 & 0.0 & 0.0 & 0.0 \\
& laptop & 95.5 & 36.3 & 4.6 & 10.2 & 11.0 \\
& mug & \underline{82.8} & 44.1 & 0.2 & 0.8 & 1.7 \\

\midrule
\multirow{5}*{Ours} & bottle & 81.9 & 51.9 & 49.7 & \textbf{94.8} & \textbf{97.2}\\
& bowl & 96.1 & 75.1 & \textbf{86.1} & \textbf{94.4} & \textbf{95.2}\\
& camera & \underline{63.0} & \underline{15.7} & \textbf{0.7} & \textbf{9.2} & \textbf{15.8}\\
& laptop & \underline{99.5} & \textbf{99.2} & \underline{13.2} & \textbf{18.1} & \underline{18.1}\\
& mug & 77.6 & 49.4 & \underline{2.0} & 9.1 & 13.1\\
\bottomrule
\end{tabular}
\caption{\textbf{Quanitative comparisons of various methods on Wild6D dataset.} The best results are \textbf{bolded} and the second-best results are \underline{underscored}.}
\label{tab:wild6d}
\end{table}

\begin{table}[!h]
\centering
\begin{tabular}{lcccccc}
\toprule
\multirow{3}*{} & \multirow{3}*{} & \multicolumn{5}{c}{AP (\%)$\uparrow$} \\
\cmidrule(lr){3-7}
~ & ~ &\multirowcell{2}{3D$_{25}$} & \multirowcell{2}{3D$_{50}$} & \multirowcell{2}{5$^\circ$\\ 5 cm} & \multirowcell{2}{10$^\circ$\\ 5 cm} & \multirowcell{2}{15$^\circ$\\ 5 cm} \\\\
\midrule

\multirow{3}*{NOCS~\cite{wang2019normalized}} & bottle & 29.2 & 0.9 & 5.3 & 30.5 & 61.4 \\ 
& can & 0.3 & 0.2 & 0.0 & 3.3 & \underline{58.0} \\
& mug & 43.4 & 5.7 & 0.0 & 0.1 & 1.0 \\
\midrule

\multirow{3}*{DualPoseNet~\cite{Lin_2021_ICCV} }& bottle & 22.6 & 0.0 & 0.9 & 13.0 & 36.6 \\ 
& can & 18.6 & 4.7 & 2.8 & 13.9 & 26.1 \\
& mug & 78.0 & 18.9 & 0.0 & 3.2 & 6.7 \\
\midrule

\multirow{3}*{SAR-Net~\cite{lin2022sar}} & bottle & \underline{85.3} & \textbf{20.2} & 0.3 & 19.8 & 54.8 \\ 
& can & 39.4 & 0.3 & 0.0 & 0.9 & 19.9 \\
& mug & 96.6 & 0.1 & 0.0 & 0.7 & 6.5 \\

\midrule
\multirow{3}*{RePoNet~\cite{ze2022category}} 
& bottle & \textbf{87.4} & \underline{18.5} & 11.0 & 60.6 & 78.0 \\
& can & - & - & - & - & -\\
& mug & \underline{97.4} & \underline{55.2} & 1.0 & 20.3 & 39.9 \\

\midrule
\multirow{3}*{HS-Pose~\cite{zheng2023hs}} & bottle & 80.7 & 9.6 & \underline{41.1} & \underline{73.9} & \underline{84.6} \\
& can & \textbf{100.0} & \underline{17.1} & \underline{6.6} & \underline{41.8} & \textbf{67.8}\\
& mug & \textbf{100.0} & 27.2 & \underline{0.6} & \underline{7.7} & \underline{12.2} \\

\midrule
\multirow{3}*{CPPF~\cite{you2022cppf}} & bottle & 15.5 & 1.5 & 13.5 & 35.0 & 41.7 \\ 
& can & \underline{88.7} & 0.3 & 3.3 & 9.4 & 10.9 \\
& mug & 96.6 & 6.7 & 0.0 & 0.0 & 0.1 \\
\midrule

\multirow{3}*{Ours} & bottle & 53.2 & 8.9 & \textbf{64.5} & \textbf{96.0} & \textbf{96.8}\\
& can & 86.8 & \textbf{25.0} & \textbf{12.7} & \textbf{46.1} & 50.0\\
& mug & \textbf{100.0} & \textbf{93.2} & \textbf{7.9} & \textbf{47.1} & \textbf{65.6}\\
\bottomrule
\end{tabular}
\caption{\textbf{Quanitative comparisons of various methods on PhoCAL dataset.} The best results are \textbf{bolded} and the second-best results are \underline{underscored}.}
\label{tab:phocal}
\end{table}

\begin{table}[!h]
\centering
\begin{tabular}{lcccccc}
\toprule
\multirow{3}*{} & \multirow{3}*{} & \multicolumn{5}{c}{AP (\%)$\uparrow$} \\
\cmidrule(lr){3-7}
~ & ~ &\multirowcell{2}{3D$_{25}$} & \multirowcell{2}{3D$_{50}$} & \multirowcell{2}{20$^\circ$\\ 5 cm} & \multirowcell{2}{40$^\circ$\\ 5 cm} & \multirowcell{2}{60$^\circ$\\ 5 cm} \\\\
\midrule
\multirow{3}*{NOCS~\cite{wang2019normalized}} & bottle & \underline{3.3} & 0.0 & 2.3 & 2.3 & 2.3 \\
& bowl & 1.7 & 0.3 & 0.5 & 0.5 & 0.5 \\
& mug & 4.9 & 2.4 & 0.0 & 1.7 & 1.9 \\
\midrule

\multirow{3}*{DualPoseNet~\cite{Lin_2021_ICCV} } & bottle & 1.4 & 0.0 & 4.3 & 9.3 & 15.8\\
& bowl & \textbf{61.3} & 2.3 & 1.4 & 16.1 & 33.3\\
& mug & \textbf{60.4} & \textbf{28.3} & 0.2 & 0.8 & 2.1\\

\midrule
\multirow{3}*{SAR-Net~\cite{lin2022sar}} & bottle & 2.5 & 0.0 & 0.1 & 6.3 & 14.2\\
& bowl & 4.8 & 0.1 & 1.2 & 28.8 & 55.6\\
& mug & 49.5 & 0.4 & 0.0 & 0.4 & 0.9\\

\midrule
\multirow{3}*{RePoNet~\cite{ze2022category}} & bottle & 1.9 & \underline{0.1} & 9.6 & \textbf{20.2} & \underline{20.2}\\
& bowl & \underline{40.3} & 2.4 & 11.5 & \underline{37.9} & \underline{66.2}\\
& mug & 21.1 & 1.4 & 0.5 & \underline{1.5} & \underline{2.3}\\

\midrule
\multirow{3}*{HS-Pose~\cite{zheng2023hs}} & bottle & 0.2 & 0.0 & 0.5 & 10.5 & 15.1 \\
& bowl & 11.2 & 0.0 & 1.4 & 10.5 & 17.3\\
& mug & 8.3 & 0.1 & 0.0 & 0.6 & 0.8 \\

\midrule
\multirow{3}*{CPPF~\cite{you2022cppf}} & bottle & 0.0 & 0.0 & \textbf{13.3} & \underline{15.5} & 15.5\\
& bowl & 32.7 & \underline{3.4} & \underline{14.5} & 37.6 & 42.1\\
& mug & 35.8 & 15.2 &\underline{0.7} & 1.0 & \underline{2.3} \\

\midrule
\multirow{3}*{Ours} & bottle & \textbf{6.9} & \textbf{0.5} & \underline{11.1} & 13.7 & \textbf{22.9}\\
& bowl & 38.5 & \textbf{15.6} & \textbf{79.5} & \textbf{92.2} & \textbf{92.2}\\
& mug & \underline{54.2} & \underline{21.2} & \textbf{22.0} & \textbf{38.9} & \textbf{47.2}\\
\bottomrule
\end{tabular}
\caption{\textbf{Quanitative comparisons of various methods on DiversePose 300 dataset.}  The best results are \textbf{bolded} and the second-best results are \underline{underscored}.}
\label{tab:diverse}
\end{table}

\subsection{Ablation Studies and More Analysis}
In this section, we validate our module design by conducting several ablation experiments. Results are given in Table~\ref{tab:ablation}.
% In addition, we also evaluate our method's performance when only bounding boxes are available.

\begin{table}[ht]
\begin{center}
% \resizebox{\linewidth}{!}{
\begin{tabular}{lccccccc}
\toprule
\multirow{3}*{} & \multirow{3}*{} & \multicolumn{5}{c}{AP (\%)$\uparrow$} \\
\cmidrule(lr){2-8}
~ & \multirowcell{2}{3D$_{25}$} & \multirowcell{2}{3D$_{50}$} & \multirowcell{2}{5$^\circ$\\ 5 cm} & \multirowcell{2}{10$^\circ$\\ 5 cm} & \multirowcell{2}{15$^\circ$\\ 5 cm} \\
& & & & & & &\\
\midrule
N = 2  & 81.5 & 53.8 & 28.7 & 63.9 & 84.4 \\
N = 10  & 82.3 & 58.0 & 32.2 & 65.6 & 83.1\\
\midrule
w/o Uncertainty & 79.0 & 50.3 & 24.2 & 62.0 & 78.0  \\
w/o Noisy Pair Filtering & 82.2 & 53.7 & 12.6 & 34.2 & 58.1 \\
w/o Point Re-weighting  & 82.3 & 54.0 & 31.0 & 65.3 & 84.9 \\
w/o Online Alignment  & 82.3 & 55.1 & 24.5  & 64.8 & 79.4 \\
\midrule
Geometric Model Only & 82.3 & 55.0 & 19.8 & 54.7 & 84.9 \\
Visual Model Only  & 75.8 & 36.9 & 22.1 & 43.5 & 56.1 \\
Concat. Geo.\&Vis. Features & 76.9 & 40.9 & 18.2 & 61.4 & 74.0 \\
\midrule
Ours & \textbf{82.4} &\textbf{ 55.2} & \textbf{32.3} & \textbf{65.9} & \textbf{85.2} \\
\bottomrule
\end{tabular}
% }
\end{center}
\caption{\textbf{Ablation studies and analysis.}}
\label{tab:ablation}
\end{table}

\subsubsection{Importance of $N$-Point Tuple.}
Table~\ref{tab:ablation} displays the results when utilizing naive point pairs ($N=2$) as opposed to $N$-point tuples. There's a noticeable decline in performance across all datasets and metrics. The data indicates that while there's a significant performance boost with increasing points, the gains tend to plateau as the number of points exceeds 5, suggesting that the information within the tuple becomes saturated.

\subsubsection{Importance of Uncertainty Modeling.}
Directly estimating voting targets, such as $\mu$ and $\nu$, leads to a marked degradation in performance. This is because the model tends to produce an arbitrary value for all input point pairs that fall within the same collision bin.

\subsubsection{Importance of Noisy Tuple Filtering.}
In real-world settings, noise is almost inevitable, stemming from factors like the precision of instance detectors and inherent noise from depth sensors. Thus, a noise-robust model like CPPF++ that can autonomously filter out background noise is crucial for effective generalization.

\subsubsection{Influence of Point Re-weighting}
Point re-weighting is instrumental in mitigating the sampling bias introduced by noisy tuple filtering. As demonstrated in Table~\ref{tab:ablation}, rebalancing the samples leads to improved results.

\subsubsection{Influence of Online Alignment Optimization}
Online alignment optimization offers a chance to correct the output, considering the voting process itself is non-differentiable. Ablation studies indicate that this strategy significantly enhances pose accuracy, particularly when errors are minimal, yielding a $7.8$ increase in AP@5$^\circ$5 cm.

\subsubsection{Choice of Tuple Features}
It is crucial to acknowledge the importance of both geometric and visual cues in determining the pose of the target object. As indicated in Table~\ref{tab:ablation}, relying solely on either geometric or visual models does not achieve optimal performance. However, merely concatenating these cues can also overwhelm our model, complicating the selection of relevant features.

\subsubsection{Training Data Preparation and Inference Time.}
% One of the advantages of our approach is that the training data can be generated pretty fast. We generate our synthetic rendering data on-the-fly throughout the training process. This efficiency is attributed to our on-the-fly training data generation without backgrounds using cost-effective non-ray-tracing methods.

The CAMERA synthetic dataset previously proposed by NOCS~\cite{wang2019normalized} (as depicted in the middle of Figure~\ref{fig:diffnocs}) necessitates the simulation of placing multiple objects on a table without collisions, a process that is significantly more time-consuming in terms of simulation and rendering. Given that NOCS has not released the generation and rendering code for the CAMERA synthetic dataset, we undertook the task of reimplementing it and observed that the rendering time for a single image is 236ms, considerably longer than the rendering time for a single object, which is 12ms. The total CAMERA synthetic dataset, comprising 300K samples, would thus require approximately 19.8 hours to render.

In contrast, our synthetic dataset comprises only a single object without any background, with the total rendering time for 200 samples across 100 epochs (equating to 20,000 samples) taking just 236.7 seconds. However, our model's incorporation of DINOv2 features necessitates their generation during the data preparation stage, currently the most significant bottleneck in our method. On average, computing these features for each sample takes 412ms. Consequently, the total data preparation time for our dataset is 236.7 seconds plus 2.3 hours, which remains faster than the rendering time for the CAMERA synthetic dataset.
% It can be trained in a mere 30 minutes 300(on a single 1080Ti) for each instance or category. This efficiency is attributed to our on-the-fly training data generation using cost-effective non-ray-tracing methods. Additionally, there's no need to wait for all votes to converge, as long as a sufficient number of accurate votes are present.

In terms of inference, our method averages at $930$ ms per object. This includes the computation of SHOT descriptors, point cloud normals and DINOv2 features. We utilize the readily available SHOT descriptor implementation from the PCL~\cite{rusu20113d} library, which averages at 207ms. The network forward of pretrained DINOv2 costs 412ms. In the future, we expect a distilled lightweight DINO-like feature extractor can be employed to boost the speed of our method.
% In the future, a tailored GPU implementation could be employed to further enhance the speed of our method.

\begin{figure*}[ht]
    \centering
    \includegraphics[width=\linewidth]{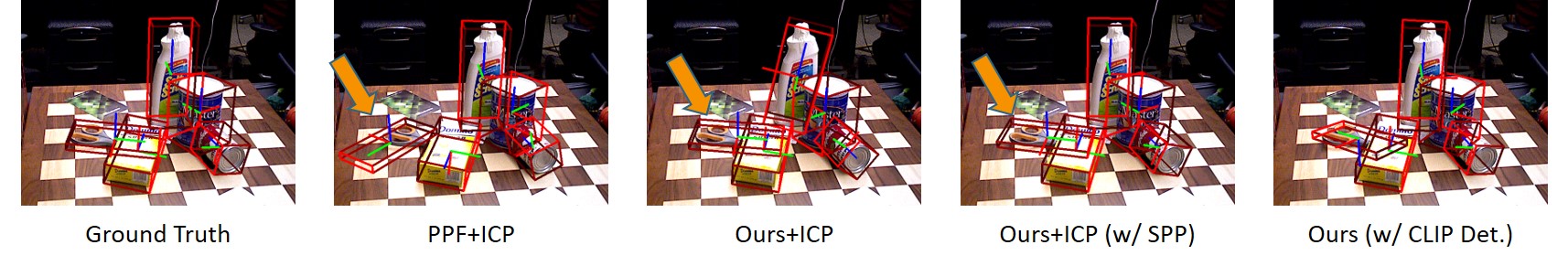}
    \caption{\textbf{Qualitative comparison on YCB-Video dataset.} Though not trained on cluttered objects, our method still gives decent results.}
    \label{fig:resycbv}
\end{figure*}

\begin{table*}[ht]
\begin{center}
\resizebox{0.85\linewidth}{!}{
\begin{tabular}{l|cc|cc|cc|cc|cc}
\toprule
% \multicolumn{1}{l|}{} & \multicolumn{6}{c|}{Syn.} & \multicolumn{4}{c}{Syn.+Real}\\
 % \cmidrule(lr){2-7}\cmidrule(lr){8-13}
 \multicolumn{1}{l|}{} & \multicolumn{2}{c|}{PPF~\cite{vidal20186d}} & \multicolumn{2}{c|}{PPF~\cite{vidal20186d}+ICP} & \multicolumn{2}{c|}{Ours} & \multicolumn{2}{c|}{Ours+ICP} & \multicolumn{2}{c}{Ours w/ CLIP Det.} \\
  \cmidrule(lr){2-3}\cmidrule(lr){4-5}\cmidrule(lr){6-7}\cmidrule(lr){8-9}\cmidrule(lr){10-11}
 Objects & ADD(-S) & ADD-S & ADD(-S) & ADD-S & ADD(-S) & ADD-S & ADD(-S) & ADD-S & ADD(-S) & ADD-S  \\
 \midrule
master\_chef\_can & 36.7 & 93.4 & 38.6 & \textbf{94.5} & 27.2 & 85.3 & \textbf{58.4} & 92.5          & 48.2          & 75.1          \\
cracker\_box      & 39.6 & 83.7 & 41.2 & 86.0          & 38.4 & 81.3 & \textbf{74.7} & \textbf{92.1} & 33.6          & 51.4          \\
sugar\_box        & 60.2 & 93.5 & 61.0 & 94.3          & 66.3 & 94.8 & \textbf{94.4} & \textbf{98.0} & 74.0          & 85.1          \\
tomato\_soup\_can & 56.8 & 94.0 & 56.9 & 94.6          & 60.4 & 93.6 & \textbf{90.2} & \textbf{94.7} & 74.8          & 77.5          \\
mustard\_bottle   & 60.3 & 91.1 & 64.9 & 92.4          & 57.5 & 88.2 & 60.8 & 90.2 & \textbf{66.0} & \textbf{95.5} \\
tuna\_fish\_can   & 62.2 & 95.6 & 63.6 & \textbf{97.2} & 62.8 & 93.8 & \textbf{72.4} & 97.0          & 11.9          & 18.5          \\
pudding\_box      & 43.8 & 91.2 & 44.8 & 92.7          & 46.9 & 84.7 & \textbf{91.4} & \textbf{96.5} & 0.0           & 0.0           \\
gelatin\_box      & 70.4 & 96.6 & 78.2 & 97.5          & 81.7 & 97.0 & \textbf{96.3} & \textbf{98.6} & 35.9          & 36.0          \\
potted\_meat\_can & 56.2 & 79.4 & 58.3 & \textbf{80.4} & 54.8 & 78.3 & \textbf{64.7} & 79.3          & 29.9          & 36.0          \\
banana            & 42.3 & 81.7 & 43.3 & 84.5          & 48.7 & 82.2 & \textbf{59.7} & \textbf{92.8} & 57.8          & 91.0          \\
pitcher\_base     & 11.1 & 26.7 & 13.0 & 27.6          & 10.4 & 27.8 & 10.7          & 25.0          & \textbf{66.7} & \textbf{91.9} \\
bleach\_cleanser  & 45.7 & 74.5 & 54.9 & 76.3          & 56.3 & 75.4 & \textbf{56.5} & 78.7          & 54.7          & \textbf{86.1} \\
bowl              & 67.7 & 67.7 & 67.9 & 67.9          & 62.4 & 62.4 & 71.0          & 71.0          & \textbf{79.3} & \textbf{79.3} \\
mug               & 39.7 & 91.6 & 40.7 & 92.0          & 41.8 & 91.6 & 48.5          & \textbf{92.8} & \textbf{59.4} & 81.5          \\
power\_drill      & 50.9 & 81.5 & 59.5 & 84.3          & 63.8 & 86.7 & \textbf{71.7} & \textbf{91.9} & 42.6          & 51.8          \\
wood\_block       & 71.4 & 71.4 & 72.0 & 72.0          & 69.1 & 69.1 & 61.7          & 61.7          & \textbf{88.1} & \textbf{88.1} \\
scissors          & 20.4 & 67.7 & 26.9 & 75.1          & 53.0 & 85.3 & \textbf{54.8} & \textbf{87.8} & 32.8          & 66.2          \\
large\_marker     & 65.1 & 89.1 & 74.7 & 95.6          & 76.3 & 95.6 & \textbf{88.1} & \textbf{96.7} & 58.4          &          63.3 \\
large\_clamp      & 79.8 & 79.8 & 83.6 & 83.6          & 84.9 & 84.9 & \textbf{88.7} & \textbf{88.7} & 9.3           &           9.3 \\
ex\_large\_clamp  & 37.1 & 37.1 & 39.2 & 39.2          & 40.4 & 40.4 & \textbf{44.6} & \textbf{44.6} & 1.1           &           1.1 \\
foam\_brick       & 89.5 & 89.5 & 90.1 & 90.1          & 88.9 & 88.9 & \textbf{90.6} & \textbf{90.6} & 46.0          &          46.0 \\
\midrule
ALL               & 52.0 & 81.6 & 55.2 & 83.4 & 55.4 & 81.8 & \textbf{70.0} & \textbf{85.5} & 46.8 & 59.3 \\
\bottomrule
\end{tabular}}
\end{center}
\caption{\textbf{AUC comparison between PPF and CPPF++ (Ours) on YCB-Video dataset.}}
\label{tab:ycbv}
\end{table*}

\subsection{Comparison with Traditional PPF}
Though our method is mainly designed for category-level pose estimation, interestingly, we find our method is superior than traditional PPF~\cite{drost2010model} in terms of instance-level pose estimation.
\subsubsection{Datasets.} We use YCB-Video~\cite{xiang2018posecnn} dataset to evaluate our method on instance-level pose estimation, where the exact 3D model of the target object is known at training time. YCB-Video dataset consists of 21 objects and 92 RGB-D video sequences with pose annotations. We follow previous work~\cite{xiang2018posecnn} to evaluate on the 2,949 keyframes in 12 videos. Objects are placed arbitrarily on the table in the test frames.

\subsubsection{Metrics.}
We follow PoseCNN~\cite{xiang2018posecnn} to use the standard ADD(-S) and ADD-S metrics and report their area-under-the-curves (AUCs). The ADD metric is first introduced in ~\cite{hinterstoisser2012model} to calculate average per-point distance between two point clouds, transformed by the predicted pose and the ground-truth, respectively. For symmetric objects like bowls, ADD-S metric is introduced to count for the point correspondence ambiguity. The notation ADD(-S) corresponds to computing ADD for non-symmetric objects and ADD-S for symmetric objects. 

\subsubsection{Baselines.}
In the context of this paper, our primary objective is to enhance performance within a stringent sim-to-real framework. Consequently, we compare our methodology with PPF~\cite{vidal20186d}, which, akin to our approach, solely requires synthetic CAD models during the training phase. Moreover, we conduct comparisons with several leading state-of-the-art category-level pose estimation methods: DualPoseNet~\cite{Lin_2021_ICCV}, SAR-Net~\cite{lin2022sar}, and HS-Pose~\cite{zheng2023hs}. These methods are retrained on the YCB-Video dataset, utilizing the same synthetic object training data as ours to ensure a fair and consistent benchmarking environment. For a fair comparison, we use the readily available instance masks from PoseCNN~\cite{xiang2018posecnn} as the detection prior. For all methods, results after applying ICP are also given. 

Given the utilization of the Iterative Closest Point (ICP) algorithm, we deactivate the online alignment optimization feature, as both methodologies aim to achieve analogous objectives. Additionally, we opt for SuperPoint~\cite{detone2018superpoint} features in place of DINO features, as the requirement shifts from needing intra-category correspondence, which DINO provides, to necessitating correspondence from the same instance, a task for which SuperPoint features are more aptly suited.

We acknowledge the emergence of recent sim-to-real techniques~\cite{haugaard2022surfemb,sundermeyer2023bop,wang2021gdr} that employ Blender for the rendering of hyper-realistic training images. Nonetheless, as highlighted in the introduction, these strategies introduce supplementary intricacies in the generation of backgrounds and disturbances. The extended duration required for data preparation and training renders them less feasible for real-world applications.

\subsubsection{Results.}
Table~\ref{tab:ycbv} presents the comprehensive results between PPF and CPPF++ for each object within the YCB-Video dataset. Our approach surpasses PPF across nearly all metrics, both with and without the ICP cases. Notably, our method does not require exposure to object occlusion or background during training and can be seamlessly transitioned to real-world scenarios. Visual examples are showcased in Figure~\ref{fig:resycbv}.

% Furthermore, to differentiate poses for objects with ambiguous geometries, such as boxes, we incorporate a SuperPoint~\cite{detone2018superpoint} based encoding technique to account for texture. This enhancement bolsters our method's capability, particularly when dealing with textured symmetric objects. Specifically, in conjunction with the tuple features elaborated in Equation~\ref{sec:npoint}, we concatenate the interpolated per-point SuperPoint feature for each tuple. This integration of color and texture information aids our model in more accurately handling symmetric geometric objects.

Moreover, we also present results utilizing the CLIP fine-tuned instance segmentation detector with Grounded FastSAM~\cite{liu2023grounding,zhao2023fast}, as detailed in Section~\ref{sec:maskpred}. While it exhibits proficiency in certain categories, its zero-shot detection performance for many objects remains suboptimal, leading to a decline in both ADD(-S) and ADD-S metrics.

Table~\ref{tab:ycb_cat} presents a comparative analysis between state-of-the-art pose estimation methods and our approach. It is evident that while these methods achieve commendable performance on category-level pose estimation datasets, they fall short when compared to our method, primarily due to their limited generalizability.

\begin{table}[ht]
    \centering
    \begin{tabular}{lcc}
    \toprule
         & ADD(-S) & ADD-S \\
        \midrule
        DualPoseNet+ICP & 15.4 & 50.2 \\
        SAR-Net+ICP & 31.6 & 70.2\\
        HS-Pose+ICP & 26.1 & 59.7 \\
        \midrule
        Ours+ICP & \textbf{70.0} & \textbf{85.5} \\
        \bottomrule
    \end{tabular}
    \caption{\textbf{AUC comparison between previous category-level pose estimation methods and ours.} Our method gives a much better result.}
    \label{tab:ycb_cat}
\end{table}
% \begin{figure}[h]
%     \centering
%     \includegraphics[width=0.7\linewidth]{figures/failure.jpg}
%     \caption{In YCB-Video dataset, the CAD model of pitcher base has a great visual difference from the images used for evaluation, resulting in a drop of our method.}
%     \label{fig:failure}
% \end{figure}

% The only failure case is the pitcher\_base object, where our method only gets 5.9 ADD(-S) and 25.4 ADD-S. The cause of this failure is the discrepancy of the provided CAD model and the test images, as shown in Figure~\ref{fig:failure}.

\subsubsection{Detailed Analysis and Failure Cases.}
\begin{figure}
    \centering
    \includegraphics[width=0.8\linewidth]{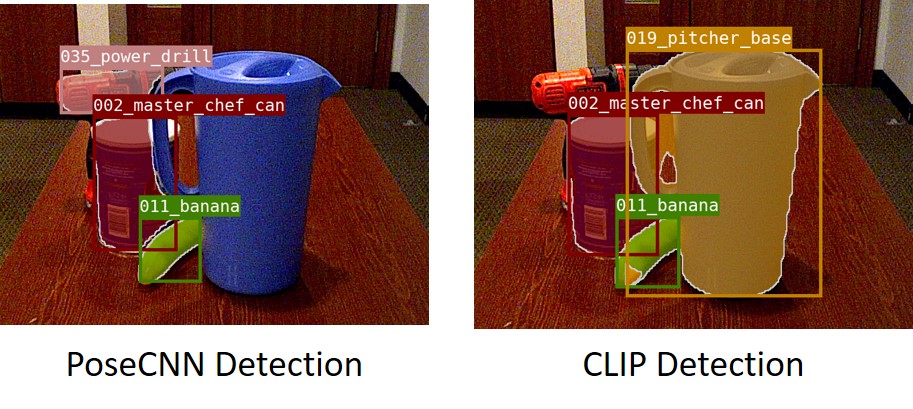} 
    \caption{\textbf{A qualitative comparison of the instance segmentation results between the pretrained model provided by PoseCNN and the CLIP finetuned model.} The PoseCNN pretrained model perform badly on the object $pitcher\_base$ while the CLIP model works well on this object. On the other hand, the CLIP finetuned model fails to detect the challenging instance $power\_drill$ due to the large occlusion.}
    \label{fig:detection}
\end{figure}

Due to the results in Table~\ref{tab:ycbv}, we find that although our method is effective in almost all the instances, there is a significant performance drop for the instance $pitcher\_base$. This is caused by the substantial failure of instance segmentation. The detector provided by PoseCNN~\cite{xiang2018posecnn} fails to detect a majority of $pitcher\_base$ instances. When we use our fine-tuned CLIP detection, the performance returns to normal. A qualitative comparison of $pitcher\_base$ detections between the PoseCNN detector and the CLIP detector is given in Figure~\ref{fig:detection}.

\section{Conclusions}
This paper presents the innovative CPPF++ method tailored for sim-to-real pose estimation. Rooted in the foundational point-pair voting scheme of CPPF, our approach reinterprets it using a probabilistic perspective. To counteract voting collision, we innovatively model voting uncertainty by gauging the probabilistic distribution of each point pair in the canonical space, complemented by an noisy tuple filtering technique to eliminate votes linked to backgrounds or clutters. Our method also introduces several useful modules is further enhance the performance: $N$-point tuple feature, online alignment optimization and tuple feature assemble. Alongside this, we unveil the DiversePose 300 dataset, curated to provide a complement assessment of top-tier methods in diverse real-world conditions. Our empirical findings validate the effectiveness of CPPF++, highlighting a marked decrease in sim-to-real gap on both category-level and instance-level datasets.

% Our method heavily depends on the depth input to predict the pose. It is more general and helpful to use RGB colors only, by exploring a novel voting strategy in the image plane.

\section{Acknowledgements}
% This work was supported by the National Key Research and Development Project of China (No. 2021ZD0110700), the National Natural Science Foundation of China under Grant 51975350, Shanghai Municipal Science and Technology Major Project (2021SHZDZX0102), Shanghai Qi Zhi Institute, and SHEITC (2018-RGZN-02046). This work was also supported by the  Shanghai AI development project (2020-RGZN-02006) and ``cross research fund for translational medicine'' of Shanghai Jiao Tong University (zh2018qnb17, zh2018qna37, YG2022ZD018).

This work was supported by the National Key Research and Development Project of China (No. 2022ZD0160102)
National Key Research and Development Project of China (No. 2021ZD0110704), Shanghai Artificial Intelligence Laboratory, XPLORER PRIZE grants. Yang You is also supported in part by the Outstanding Doctoral Graduates Development Scholarship of Shanghai Jiao Tong University.

%%%%%%%%% REFERENCES
{\small
\bibliographystyle{ieee_fullname}
\bibliography{egbib}

\begin{thebibliography}{10}\itemsep=-1pt

\bibitem{ahmadyan2021objectron}
Adel Ahmadyan, Liangkai Zhang, Artsiom Ablavatski, Jianing Wei, and Matthias
  Grundmann.
\newblock Objectron: A large scale dataset of object-centric videos in the wild
  with pose annotations.
\newblock In {\em Proceedings of the IEEE/CVF Conference on Computer Vision and
  Pattern Recognition}, pages 7822--7831, 2021.

\bibitem{shapenet2015}
Angel~X. Chang, Thomas Funkhouser, Leonidas Guibas, Pat Hanrahan, Qixing Huang,
  Zimo Li, Silvio Savarese, Manolis Savva, Shuran Song, Hao Su, Jianxiong Xiao,
  Li Yi, and Fisher Yu.
\newblock {ShapeNet: An Information-Rich 3D Model Repository}.
\newblock Technical Report arXiv:1512.03012 [cs.GR], Stanford University ---
  Princeton University --- Toyota Technological Institute at Chicago, 2015.

\bibitem{chen2020learning}
Dengsheng Chen, Jun Li, Zheng Wang, and Kai Xu.
\newblock Learning canonical shape space for category-level 6d object pose and
  size estimation.
\newblock In {\em Proceedings of the IEEE/CVF conference on computer vision and
  pattern recognition}, pages 11973--11982, 2020.

\bibitem{chen2021fs}
Wei Chen, Xi Jia, Hyung~Jin Chang, Jinming Duan, Linlin Shen, and Ales
  Leonardis.
\newblock Fs-net: Fast shape-based network for category-level 6d object pose
  estimation with decoupled rotation mechanism.
\newblock In {\em Proceedings of the IEEE/CVF Conference on Computer Vision and
  Pattern Recognition}, pages 1581--1590, 2021.

\bibitem{chen2020category}
Xu Chen, Zijian Dong, Jie Song, Andreas Geiger, and Otmar Hilliges.
\newblock Category level object pose estimation via neural
  analysis-by-synthesis.
\newblock In {\em European Conference on Computer Vision}, pages 139--156.
  Springer, 2020.

\bibitem{deng2020self}
Xinke Deng, Yu Xiang, Arsalan Mousavian, Clemens Eppner, Timothy Bretl, and
  Dieter Fox.
\newblock Self-supervised 6d object pose estimation for robot manipulation.
\newblock In {\em 2020 IEEE International Conference on Robotics and Automation
  (ICRA)}, pages 3665--3671. IEEE, 2020.

\bibitem{detone2018superpoint}
Daniel DeTone, Tomasz Malisiewicz, and Andrew Rabinovich.
\newblock Superpoint: Self-supervised interest point detection and description.
\newblock In {\em Proceedings of the IEEE conference on computer vision and
  pattern recognition workshops}, pages 224--236, 2018.

\bibitem{drost2010model}
Bertram Drost, Markus Ulrich, Nassir Navab, and Slobodan Ilic.
\newblock Model globally, match locally: Efficient and robust 3d object
  recognition.
\newblock In {\em 2010 IEEE computer society conference on computer vision and
  pattern recognition}, pages 998--1005. Ieee, 2010.

\bibitem{gao20206d}
Ge Gao, Mikko Lauri, Yulong Wang, Xiaolin Hu, Jianwei Zhang, and Simone
  Frintrop.
\newblock 6d object pose regression via supervised learning on point clouds.
\newblock In {\em 2020 IEEE International Conference on Robotics and Automation
  (ICRA)}, pages 3643--3649. IEEE, 2020.

\bibitem{haugaard2022surfemb}
Rasmus~Laurvig Haugaard and Anders~Glent Buch.
\newblock Surfemb: Dense and continuous correspondence distributions for object
  pose estimation with learnt surface embeddings.
\newblock In {\em Proceedings of the IEEE/CVF Conference on Computer Vision and
  Pattern Recognition}, pages 6749--6758, 2022.

\bibitem{he2017mask}
Kaiming He, Georgia Gkioxari, Piotr Doll{\'a}r, and Ross Girshick.
\newblock Mask r-cnn.
\newblock In {\em Proceedings of the IEEE international conference on computer
  vision}, pages 2961--2969, 2017.

\bibitem{hinterstoisser2012model}
Stefan Hinterstoisser, Vincent Lepetit, Slobodan Ilic, Stefan Holzer, Gary
  Bradski, Kurt Konolige, and Nassir Navab.
\newblock Model based training, detection and pose estimation of texture-less
  3d objects in heavily cluttered scenes.
\newblock In {\em Asian conference on computer vision}, pages 548--562.
  Springer, 2012.

\bibitem{hodavn2020bop}
Tom{\'a}{\v{s}} Hoda{\v{n}}, Martin Sundermeyer, Bertram Drost, Yann Labb{\'e},
  Eric Brachmann, Frank Michel, Carsten Rother, and Ji{\v{r}}{\'\i} Matas.
\newblock Bop challenge 2020 on 6d object localization.
\newblock In {\em European Conference on Computer Vision}, pages 577--594.
  Springer, 2020.

\bibitem{irshad2022centersnap}
Muhammad~Zubair Irshad, Thomas Kollar, Michael Laskey, Kevin Stone, and Zsolt
  Kira.
\newblock Centersnap: Single-shot multi-object 3d shape reconstruction and
  categorical 6d pose and size estimation.
\newblock {\em arXiv preprint arXiv:2203.01929}, 2022.

\bibitem{irshad2022shapo}
Muhammad~Zubair Irshad, Sergey Zakharov, Rares Ambrus, Thomas Kollar, Zsolt
  Kira, and Adrien Gaidon.
\newblock Shapo: Implicit representations for multi-object shape, appearance,
  and pose optimization.
\newblock {\em arXiv preprint arXiv:2207.13691}, 2022.

\bibitem{kirillov2023segment}
Alexander Kirillov, Eric Mintun, Nikhila Ravi, Hanzi Mao, Chloe Rolland, Laura
  Gustafson, Tete Xiao, Spencer Whitehead, Alexander~C Berg, Wan-Yen Lo, et~al.
\newblock Segment anything.
\newblock {\em arXiv preprint arXiv:2304.02643}, 2023.

\bibitem{labbe2020cosypose}
Yann Labb{\'e}, Justin Carpentier, Mathieu Aubry, and Josef Sivic.
\newblock Cosypose: Consistent multi-view multi-object 6d pose estimation.
\newblock In {\em European Conference on Computer Vision}, pages 574--591.
  Springer, 2020.

\bibitem{lee2022uda}
Taeyeop Lee, Byeong-Uk Lee, Inkyu Shin, Jaesung Choe, Ukcheol Shin, In~So
  Kweon, and Kuk-Jin Yoon.
\newblock Uda-cope: unsupervised domain adaptation for category-level object
  pose estimation.
\newblock In {\em Proceedings of the IEEE/CVF Conference on Computer Vision and
  Pattern Recognition}, pages 14891--14900, 2022.

\bibitem{lee2023tta}
Taeyeop Lee, Jonathan Tremblay, Valts Blukis, Bowen Wen, Byeong-Uk Lee, Inkyu
  Shin, Stan Birchfield, In~So Kweon, and Kuk-Jin Yoon.
\newblock Tta-cope: Test-time adaptation for category-level object pose
  estimation.
\newblock In {\em Proceedings of the IEEE/CVF Conference on Computer Vision and
  Pattern Recognition}, pages 21285--21295, 2023.

\bibitem{li2023mask}
Feng Li, Hao Zhang, Huaizhe Xu, Shilong Liu, Lei Zhang, Lionel~M Ni, and
  Heung-Yeung Shum.
\newblock Mask dino: Towards a unified transformer-based framework for object
  detection and segmentation.
\newblock In {\em Proceedings of the IEEE/CVF Conference on Computer Vision and
  Pattern Recognition}, pages 3041--3050, 2023.

\bibitem{li2018deepim}
Yi Li, Gu Wang, Xiangyang Ji, Yu Xiang, and Dieter Fox.
\newblock Deepim: Deep iterative matching for 6d pose estimation.
\newblock In {\em Proceedings of the European Conference on Computer Vision
  (ECCV)}, pages 683--698, 2018.

\bibitem{lin2022sar}
Haitao Lin, Zichang Liu, Chilam Cheang, Yanwei Fu, Guodong Guo, and Xiangyang
  Xue.
\newblock Sar-net: Shape alignment and recovery network for category-level 6d
  object pose and size estimation.
\newblock In {\em Proceedings of the IEEE/CVF Conference on Computer Vision and
  Pattern Recognition}, pages 6707--6717, 2022.

\bibitem{Lin_2021_ICCV}
Jiehong Lin, Zewei Wei, Zhihao Li, Songcen Xu, Kui Jia, and Yuanqing Li.
\newblock Dualposenet: Category-level 6d object pose and size estimation using
  dual pose network with refined learning of pose consistency.
\newblock In {\em Proceedings of the IEEE/CVF International Conference on
  Computer Vision (ICCV)}, pages 3560--3569, October 2021.

\bibitem{liu2023grounding}
Shilong Liu, Zhaoyang Zeng, Tianhe Ren, Feng Li, Hao Zhang, Jie Yang, Chunyuan
  Li, Jianwei Yang, Hang Su, Jun Zhu, et~al.
\newblock Grounding dino: Marrying dino with grounded pre-training for open-set
  object detection.
\newblock {\em arXiv preprint arXiv:2303.05499}, 2023.

\bibitem{matl2018}
Matthew Matl.
\newblock Pyrender.
\newblock \url{https://github.com/mmatl/pyrender}, 2018.

\bibitem{nguyen2022templates}
Van~Nguyen Nguyen, Yinlin Hu, Yang Xiao, Mathieu Salzmann, and Vincent Lepetit.
\newblock Templates for 3d object pose estimation revisited: Generalization to
  new objects and robustness to occlusions.
\newblock In {\em Proceedings of the IEEE/CVF Conference on Computer Vision and
  Pattern Recognition}, pages 6771--6780, 2022.

\bibitem{oquab2023dinov2}
Maxime Oquab, Timoth{\'e}e Darcet, Th{\'e}o Moutakanni, Huy Vo, Marc
  Szafraniec, Vasil Khalidov, Pierre Fernandez, Daniel Haziza, Francisco Massa,
  Alaaeldin El-Nouby, et~al.
\newblock Dinov2: Learning robust visual features without supervision.
\newblock {\em arXiv preprint arXiv:2304.07193}, 2023.

\bibitem{radford2021learning}
Alec Radford, Jong~Wook Kim, Chris Hallacy, Aditya Ramesh, Gabriel Goh,
  Sandhini Agarwal, Girish Sastry, Amanda Askell, Pamela Mishkin, Jack Clark,
  et~al.
\newblock Learning transferable visual models from natural language
  supervision.
\newblock In {\em International conference on machine learning}, pages
  8748--8763. PMLR, 2021.

\bibitem{rusu20113d}
Radu~Bogdan Rusu and Steve Cousins.
\newblock 3d is here: Point cloud library (pcl).
\newblock In {\em 2011 IEEE international conference on robotics and
  automation}, pages 1--4. IEEE, 2011.

\bibitem{Sager_2022}
Christoph Sager, Patrick Zschech, and Niklas Kuhl.
\newblock {labelCloud}: A lightweight labeling tool for domain-agnostic 3d
  object detection in point clouds.
\newblock {\em Computer-Aided Design and Applications}, 19(6):1191--1206, mar
  2022.

\bibitem{salti2014shot}
Samuele Salti, Federico Tombari, and Luigi Di~Stefano.
\newblock Shot: Unique signatures of histograms for surface and texture
  description.
\newblock {\em Computer Vision and Image Understanding}, 125:251--264, 2014.

\bibitem{stevvsic2020learning}
Stefan Stev{\v{s}}i{\'c}, Sammy Christen, and Otmar Hilliges.
\newblock Learning to assemble: Estimating 6d poses for robotic object-object
  manipulation.
\newblock {\em IEEE Robotics and Automation Letters}, 5(2):1159--1166, 2020.

\bibitem{su2019deep}
Yongzhi Su, Jason Rambach, Nareg Minaskan, Paul Lesur, Alain Pagani, and Didier
  Stricker.
\newblock Deep multi-state object pose estimation for augmented reality
  assembly.
\newblock In {\em 2019 IEEE International Symposium on Mixed and Augmented
  Reality Adjunct (ISMAR-Adjunct)}, pages 222--227. IEEE, 2019.

\bibitem{sundermeyer2023bop}
Martin Sundermeyer, Tom{\'a}{\v{s}} Hoda{\v{n}}, Yann Labbe, Gu Wang, Eric
  Brachmann, Bertram Drost, Carsten Rother, and Ji{\v{r}}{\'\i} Matas.
\newblock Bop challenge 2022 on detection, segmentation and pose estimation of
  specific rigid objects.
\newblock In {\em Proceedings of the IEEE/CVF Conference on Computer Vision and
  Pattern Recognition}, pages 2784--2793, 2023.

\bibitem{teed2021tangent}
Zachary Teed and Jia Deng.
\newblock Tangent space backpropagation for 3d transformation groups.
\newblock In {\em Proceedings of the IEEE/CVF Conference on Computer Vision and
  Pattern Recognition (CVPR)}, 2021.

\bibitem{tian2020shape}
Meng Tian, Marcelo~H Ang, and Gim~Hee Lee.
\newblock Shape prior deformation for categorical 6d object pose and size
  estimation.
\newblock In {\em European Conference on Computer Vision}, pages 530--546.
  Springer, 2020.

\bibitem{vidal20186d}
Joel Vidal, Chyi-Yeu Lin, and Robert Mart{\'\i}.
\newblock 6d pose estimation using an improved method based on point pair
  features.
\newblock In {\em 2018 4th international conference on control, automation and
  robotics (iccar)}, pages 405--409. IEEE, 2018.

\bibitem{wang2019densefusion}
Chen Wang, Danfei Xu, Yuke Zhu, Roberto Mart{\'\i}n-Mart{\'\i}n, Cewu Lu, Li
  Fei-Fei, and Silvio Savarese.
\newblock Densefusion: 6d object pose estimation by iterative dense fusion.
\newblock In {\em Proceedings of the IEEE/CVF conference on computer vision and
  pattern recognition}, pages 3343--3352, 2019.

\bibitem{wang2021gdr}
Gu Wang, Fabian Manhardt, Federico Tombari, and Xiangyang Ji.
\newblock Gdr-net: Geometry-guided direct regression network for monocular 6d
  object pose estimation.
\newblock In {\em Proceedings of the IEEE/CVF Conference on Computer Vision and
  Pattern Recognition}, pages 16611--16621, 2021.

\bibitem{wang2019normalized}
He Wang, Srinath Sridhar, Jingwei Huang, Julien Valentin, Shuran Song, and
  Leonidas~J Guibas.
\newblock Normalized object coordinate space for category-level 6d object pose
  and size estimation.
\newblock In {\em Proceedings of the IEEE/CVF Conference on Computer Vision and
  Pattern Recognition}, pages 2642--2651, 2019.

\bibitem{wang2022phocal}
Pengyuan Wang, HyunJun Jung, Yitong Li, Siyuan Shen, Rahul~Parthasarathy
  Srikanth, Lorenzo Garattoni, Sven Meier, Nassir Navab, and Benjamin Busam.
\newblock Phocal: A multi-modal dataset for category-level object pose
  estimation with photometrically challenging objects.
\newblock In {\em Proceedings of the IEEE/CVF conference on computer vision and
  pattern recognition}, pages 21222--21231, 2022.

\bibitem{wu2022vote}
Yangzheng Wu, Mohsen Zand, Ali Etemad, and Michael Greenspan.
\newblock Vote from the center: 6 dof pose estimation in rgb-d images by radial
  keypoint voting.
\newblock In {\em European Conference on Computer Vision (ECCV)}. Springer,
  2022.

\bibitem{xiang2018posecnn}
Yu Xiang, Tanner Schmidt, Venkatraman Narayanan, and Dieter Fox.
\newblock Posecnn: A convolutional neural network for 6d object pose estimation
  in cluttered scenes.
\newblock 2018.

\bibitem{you2022cppf}
Yang You, Ruoxi Shi, Weiming Wang, and Cewu Lu.
\newblock Cppf: Towards robust category-level 9d pose estimation in the wild.
\newblock In {\em Proceedings of the IEEE/CVF Conference on Computer Vision and
  Pattern Recognition}, pages 6866--6875, 2022.

\bibitem{ze2022category}
Yanjie Ze and Xiaolong Wang.
\newblock Category-level 6d object pose estimation in the wild: A
  semi-supervised learning approach and a new dataset.
\newblock {\em Advances in Neural Information Processing Systems},
  35:27469--27483, 2022.

\bibitem{zhao2023fast}
Xu Zhao, Wenchao Ding, Yongqi An, Yinglong Du, Tao Yu, Min Li, Ming Tang, and
  Jinqiao Wang.
\newblock Fast segment anything.
\newblock {\em arXiv preprint arXiv:2306.12156}, 2023.

\bibitem{zheng2023hs}
Linfang Zheng, Chen Wang, Yinghan Sun, Esha Dasgupta, Hua Chen, Ale{\v{s}}
  Leonardis, Wei Zhang, and Hyung~Jin Chang.
\newblock Hs-pose: Hybrid scope feature extraction for category-level object
  pose estimation.
\newblock In {\em Proceedings of the IEEE/CVF Conference on Computer Vision and
  Pattern Recognition}, pages 17163--17173, 2023.

\bibitem{zhong2022sim2real}
Chengliang Zhong, Chao Yang, Fuchun Sun, Jinshan Qi, Xiaodong Mu, Huaping Liu,
  and Wenbing Huang.
\newblock Sim2real object-centric keypoint detection and description.
\newblock In {\em Proceedings of the AAAI Conference on Artificial
  Intelligence}, volume~36, pages 5440--5449, 2022.

\end{thebibliography}
}

\end{document}